\documentclass[times,twocolumn,final,authoryear]{article}

\usepackage{framed,multirow}

\usepackage{amssymb}
\usepackage{latexsym}

\usepackage{url}
\usepackage{xcolor}
\definecolor{newcolor}{rgb}{.8,.349,.1}
\usepackage[round]{natbib}

\usepackage{algorithm} 
\usepackage{algpseudocode} 

\usepackage{array}

\usepackage[caption=false,font=footnotesize]{subfig}

\usepackage{stfloats}
\usepackage{url}

\usepackage{rotating}
\usepackage{multirow}
\usepackage{soul}
\usepackage{tikz}
\usepackage{scalefnt}

\RequirePackage{geometry}
\begin{document}


\title{Adapted and Oversegmenting Graphs: Application to Geometric Deep Learning}

\author{
  LastName1, FirstName1\\
  \texttt{first1.last1@xxxxx.com}
  \and
  LastName2, FirstName2\\
  \texttt{first2.last2@xxxxx.com}
}

\author{Alberto Gomez $^1$\\
	\and Veronika~A.~{Zimmer} $^1$\\
	\and Bishesh Khanal $^{1,2}$\\
	\and Robert {Wright} $^1$\\
	\and Nicolas {Toussaint} $^1$\\
	\and Julia~A.~{Schnabel}$^1$}

\date{%
    $^1$Department of Biomedical Engineering, School of Biomedical Engineering \& Imaging Sciences, King's College London\\%
    $^2$NAAMII, Kathmandu, Nepal\\[2ex]%
}

%


\maketitle

\begin{abstract}
We propose a novel iterative method to adapt a a graph to $d$-dimensional image data. The method drives the nodes of the graph towards image features. The adaptation process naturally lends itself to a measure of feature saliency which can then be used to retain meaningful nodes and edges in the graph. From the adapted graph, we also propose the computation of a dual graph, which inherits the saliency measure from the adapted graph, and whose edges run along image features, hence producing an oversegmenting graph. 
The proposed method is computationally efficient and fully parallelisable. We propose two distance measures to find image saliency along graph edges, and evaluate the performance on synthetic images and on natural images from publicly available databases. In both cases, the most salient nodes of the graph achieve average boundary recall over $90\%$. We also apply our method to image classification on the MNIST hand-written digit dataset, using a recently proposed Deep Geometric Learning architecture, and achieving state-of-the-art classification accuracy, for a graph-based method, of 97.86\%.
\end{abstract}

Image representation,  Graph theory, Deep Geometric Learning.


\section{Introduction}\label{sec:introduction}

High level information contained in images can be generally described using a sparse set of features. These features, encoded into \emph{feature descriptors} \citep{Lowe2004,Bay2008Speeded-UpSURF}, can then be used for different computer vision tasks, for example to perform image classification \citep{Csurka2004VisualKeypoints,Jegou2010AggregatingRepresentation,kipf2016semi,defferrard2016convolutional}.

These techniques rely on the detection and characterization of features in images. The vast majority of the literature in the field focuses on the localization of feature points in space or in space-scale \citep{kadir2001saliency,Lowe2004,Chatfield2011TheMethods,YongzhenHuang2014FeatureStudy}, however there is little published work on establishing meaningful topological relations between the detected feature points, and more generally on \emph{feature topology} in images. One reason for this is that many computer vision problems can be solved without this knowledge. For example, in image classification, the bag of visual words paradigm \citep{Csurka2004VisualKeypoints} uses feature histograms which do not consider spatial arrangement of the feature points. Moreover, salient point detection methods are normally a pre-processing step and therefore yield an over-detection of salient points, which is subsequently refined. Efforts to incorporate spatial information in computer vision, are commonly based on spatial pyramids \citep{LazebnikBeyondCategories,YangqingJia2012BeyondFeatures} (which consist on  hierarchical refinement of regular grids to achieve spatial localization of feature points) and are used to provide structurally-discriminant image representations. However these techniques do not explicitly capture image structure.

\begin{figure}[!htb]
\centering
\subfloat[Initial graph\label{fig:example1}]{
  \begin{tikzpicture}
      \node[anchor=south west,inner sep=0] (image) at (0,0) {\includegraphics[width=0.49\linewidth]{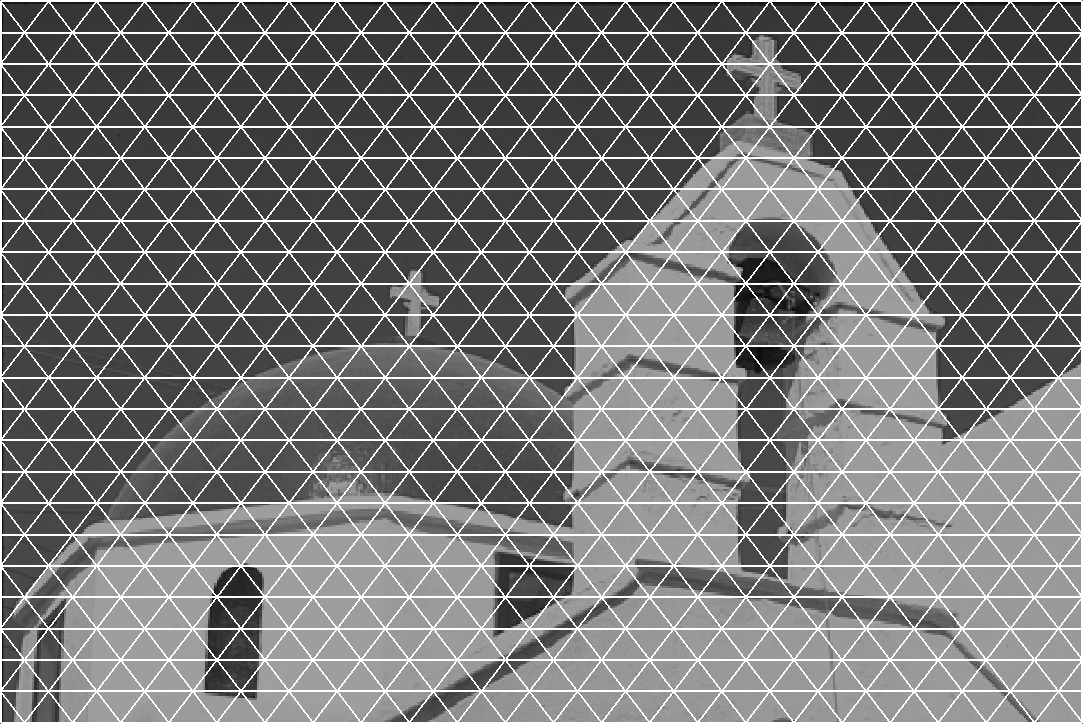}};
       \begin{scope}[x={(image.south east)},y={(image.north west)}]
          \draw[white,ultra thick] (0.3,0.45) rectangle (0.65,0.75);
      \end{scope}
  \end{tikzpicture}
}
\subfloat[Initial graph (detail)\label{fig:example2}]{
    \includegraphics[width=0.49\linewidth,trim={350pt 320pt 350pt 150pt},clip]{figures/initialGraph}
}

\subfloat[Adapted graph (detail)\label{fig:example3}]{
    \includegraphics[width=0.49\linewidth,trim={350pt 320pt 350pt 150pt},clip]{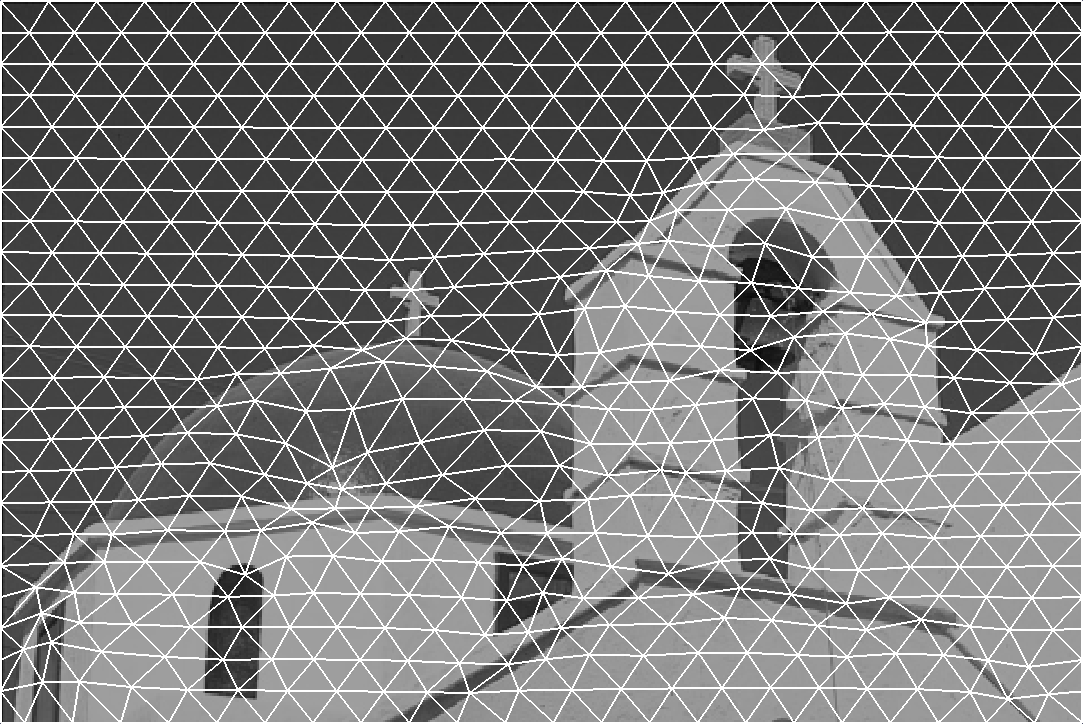}
}
\subfloat[Salient points (detail)\label{fig:example4}]{
    \includegraphics[width=0.49\linewidth,trim={350pt 320pt 350pt 150pt},clip]{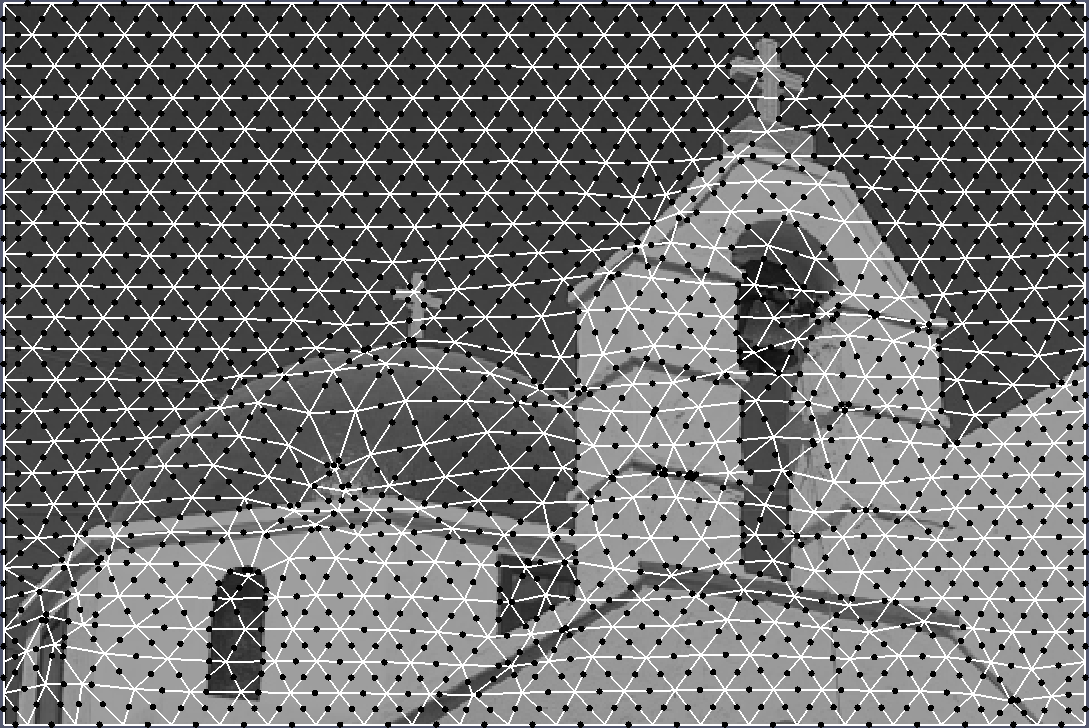}
}

\subfloat[Dual graph (detail)\label{fig:example5}]{
    \includegraphics[width=0.49\linewidth,trim={350pt 320pt 350pt 150pt},clip]{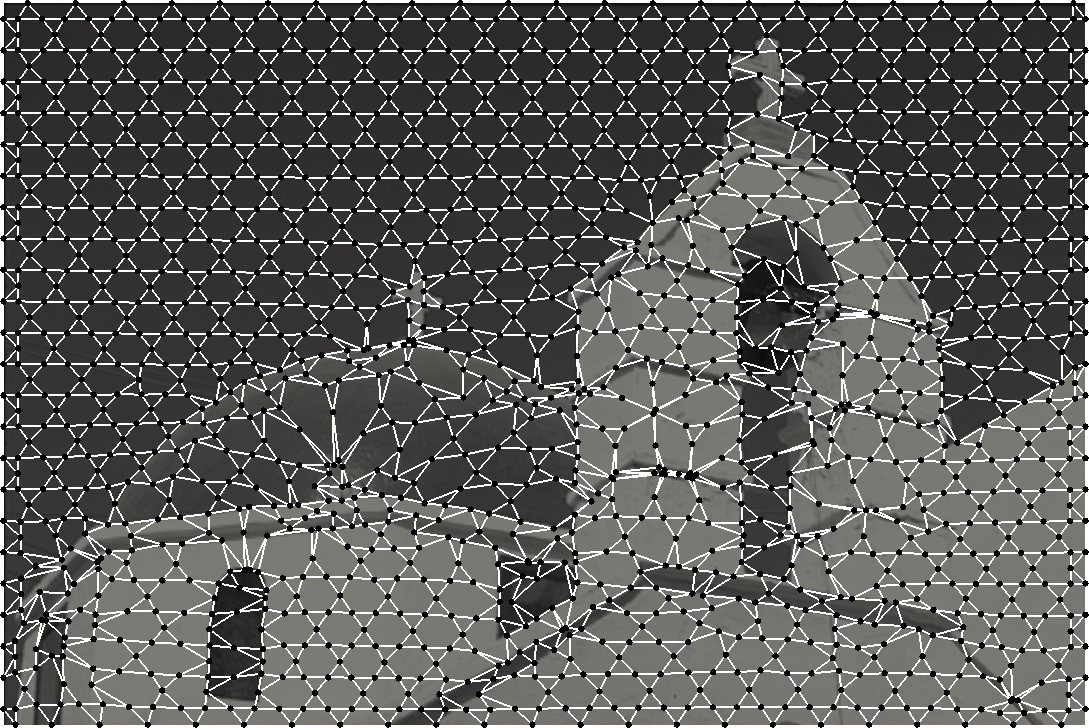}
}
\subfloat[Saliency-coded dual graph\label{fig:example6}]{
    \begin{tikzpicture}
      \node[anchor=south west,inner sep=0] (image) at (0,0) {\includegraphics[width=0.49\linewidth]{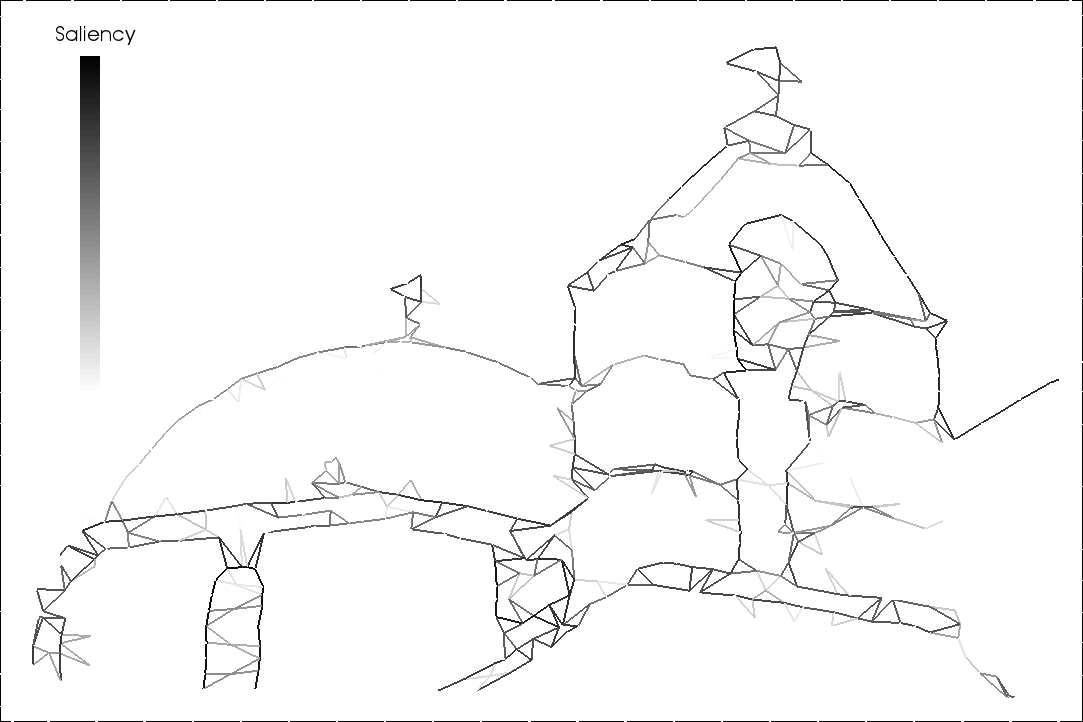}};
       \begin{scope}[x={(image.south east)},y={(image.north west)}]
          \draw[gray,ultra thick] (0.3,0.45) rectangle (0.65,0.75);
      \end{scope}
  \end{tikzpicture}
}
\caption{Summary of the proposed oversegmenting graph method. Figure \ref{fig:example1} shows an input image taken from the BSDS300 database. The rectangle indicates the zoomed region that is shown in the other figures for better appreciation. Initially, a uniform triangulated graph is placed in the image (Fig. \ref{fig:example2}). This graph is adapted to the image so that edges intersect with image boundaries (Fig. \ref{fig:example3}) at salient points, indicated as black dots in Fig. \ref{fig:example4}. To obtain an oversegmenting graph, where edges run along image boundaries, we obtain the dual of the adapted graph by connecting salient points (Fig. \ref{fig:example5}). The result, coded by saliency in grayscale (white is no salient, black is most salient), is shown in Fig. \ref{fig:example6}.  \label{fig:example}} 
\end{figure}

To compensate for the lack of feature topology, context-aware feature point  descriptors have been proposed \citep{Heinrich2012MIND:Registration,Rivaz2014Self-similarityMetric,Jiang2016MiLBP:Registration}. These descriptors can capture the appearance of a feature in relation to its surroundings, which has been shown to increase accuracy in image matching; still they cannot establish the relation in space with nearby feature points.

Recently, machine learning methods on graphs, and specifically Geometric Deep Learning \citep{ haussler1999convolution,bronstein2017geometric}, have gained interest in the community. These techniques normally act on data which are intrinsically represented in the form of a graph, for example social networks or meshed surfaces. Digital images can also be considered as graphs, where pixel centres are the nodes, intensity values are node features, and edges between nodes are established between neighbouring pixels \citep{grady2006random}. However this is not a particularly useful graph representation of an image, since the graph topology and features do not add to or synthesise the information available in the image already. 

A more interesting approach is to build a graph with fewer nodes than the number of pixels, where the nodes are salient points in the image for some definition of saliency \citep{brun2018trends}. Most related methods build graphs from a set of feature points (detected with any of the methods mentioned above) and then compute edges from feature points by Delaunay triangulation \citep{Zhou2016FactorizedMatching}, or by $k$-nearest neighbours \citep{Aguilar2009}. A related technique, image meshing, aims at extracting an (ideally) uniform triangular mesh from an image, so that image contours are aligned with triangle edges, normally in the aim of using the resulting mesh as a discretized domain for computational modelling using the Finite Element Method (FEM) \citep{Boyd2006SmoothData,Si2015TetGenGenerator}, for example to simulate flow, mechanics, or other physical phenomena. Although this normally requires a segmentation step, a method to carry out the image meshing in one step, called \emph{image-based variational meshing } was proposed by \cite{Goksel2011Image-BasedMeshing}. This method adapts an entire triangular mesh to an image, by defining an objective function related to the goodness-of-fit of the mesh to the image data and searching for the optimum of such function, at a relatively high computational cost. These approaches have two main drawbacks: first the added computational cost of computing the edges after the feature points have been detected; and second, there is no mechanism enforcing that edges have a meaningful relation with the underlying structure or with true relations between feature points other than spatial proximity. Also starting from an initial grid, another approach is to move connected grid points to local intensity maxima \citep{tuytelaars2010dense,gomez2017fast}, or to use grid nodes as seeds for a superpixel oversegmentation of the image \citep{defferrard2016convolutional}. The latter method has been recently used to demonstrate the use of geometric deep learning to classify images using superpixel-based graphs computed on them by \cite{defferrard2016convolutional,kipf2016semi,monti2017geometric,fey2018splinecnn}. 

All the above mentioned approaches that use graph representations assume that the information needed for matching is embedded in feature descriptors at the nodes and that spatial relations between these descriptors (encoded into edges) are supplementary or just computationally convenient. An exception to the above approaches, for the specific application of shape classification using binary images, is the construction of graphs from image skeletons \citep{DiRuberto2004RecognitionGraphs,XiangBai2008PathMatching,Baseski2009DissimilarityContext,battistone2018tglstm,saha2016survey}. The resulting graphs, which run along the medial shape line, are descriptive of the structure of the binary object, however by construction nodes are far from edges, corners or more generally any salient feature points, rendering this class of approaches not very well suitable for establishing correspondences or more generally for describing images through features.

In this paper, we propose a novel method to extract a graph that adapts to image features, named `\emph{adapted graph}', from which we derive image structure as a connected graph, named  `\emph{oversegmenting graph}'. The nodes of the  oversegmenting graph are located at salient image points, the edges approximate an over-segmentation of the image as the graph resolution increases, and a saliency measure is associated with each edge. Both graphs provide a sparse topological representation of image data where both nodes and edges carry meaningful information of the image structure and the computation is reduced to a one-dimensional problem, hence they can be computed very efficiently. Using very recent deep geometric learning architectures, we demonstrate the representation power of our proposed graphs on an image classification task.

The remainder of the paper is organised as follows. The proposed method is described in Sec. \ref{sec:methods}, including the salient point detection (Sec. \ref{sec:salientdetection}), the iterative adaptation process to obtain the adapted graph (Sec. \ref{sec:iterativeadaptation}) and the computation of the oversegmenting graph (Sec. \ref{sec:dualgraphs}). The experiments are described in Sec. \ref{sec:experiments} and the results, including an evaluation on the boundary adherence of the oversegmenting graphs compared to superpixels, and the performance of the adapted graphs for image classification using geometric deep learning, are provided in Sec. \ref{sec:results}.   

\section{Methods}
\label{sec:methods}
The proposed method is generic for $d$-dimensional ($d$-D) image data. For simplicity, it will be presented for 2D images, and precisions on other dimensions will be made in the Discussion.

In short, an initial graph is iteratively adapted to image features, computing a saliency measure for each feature along the way. Then the oversegmenting graph is computed as the dual of the adapted graph. This process is illustrated in Fig. \ref{fig:example}, and described more formally in Sec. \ref{sec:salientdetection}, \ref{sec:saliencyMeasure} and \ref{sec:dualgraphs} below.

\subsection{Salient Point Detection}
\label{sec:salientdetection}

Let $\mathcal{G}=(E,V)$ be a graph where $V = \{\mathbf{x}_i\}_{i=1,...,N_V}, \mathbf{x}_i \in \mathbb{R}^d$ is a set of $N_V$ $d$-dimensional nodes, and $E = \{e_i\}_{i=1,...,N_E}, e_i \in \{1,...,N_V\} \times \{1,...,N_V\}$ is a set of edges between the nodes. Initially, $G$ is initialized as a homogeneous triangulated mesh that covers the image extent, as shown in Fig. \ref{fig:example1}. For each edge $e_i=(j,k)$, we first search for the point $\hat{ \mathbf{x} } \in \mathbb{R}^d$ between the two end points $\mathbf{x}_j$ and $\mathbf{x}_k$ that is on an image feature point. 
We propose two methods to find $\hat{ \mathbf{x} }$, described in Sec. \ref{sec:distancebased} and \ref{sec:integrateddistance} respectively.

\begin{figure}[htb!]
\captionsetup[subfigure]{labelformat=empty}
	\centering
     \subfloat{
    \begin{minipage}[c]{0.05\linewidth}
    \rotatebox[origin=r]{90}{\shortstack[l]{Inputs}}
    \end{minipage}
    }
    \subfloat[Clean]{
	\raisebox{-.5\height}{\includegraphics[width=0.15\linewidth]{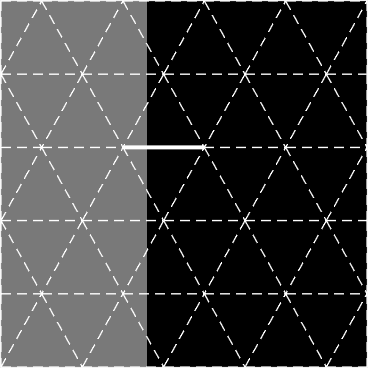}}
	}
    \subfloat[$\sigma=40\%$]{
	\raisebox{-.5\height}{\includegraphics[width=0.15\linewidth]{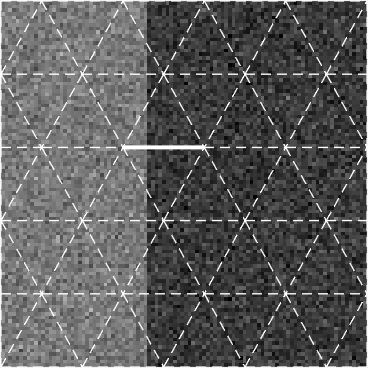}}
    }
     \subfloat[$\sigma = 60\%$]{
	\raisebox{-.5\height}{\includegraphics[width=0.15\linewidth]{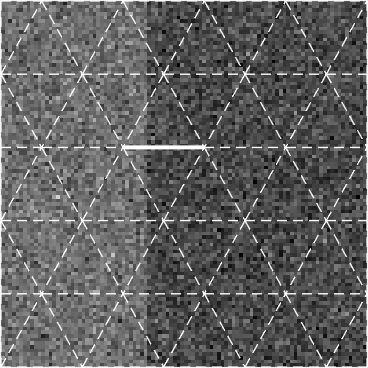}}
    }
    
    
    \subfloat[SLIC distance -- clean]{
    \raisebox{-.5\height}{\includegraphics[width=0.48\linewidth]{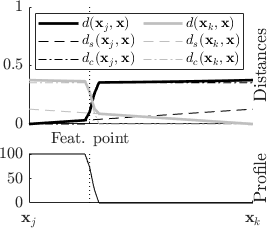}}
    }
    \subfloat[Robust saliency -- clean]{
    \raisebox{-.5\height}{\includegraphics[width=0.48\linewidth]{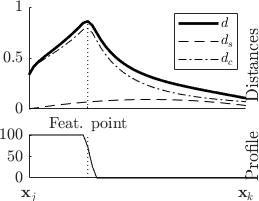}}
    }
    
    \subfloat[SLIC distance -- $\sigma = 30\%$]{
    \raisebox{-.5\height}{\includegraphics[width=0.48\linewidth]{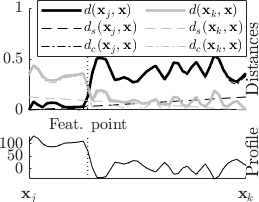}}
    }
    \subfloat[Robust saliency -- $\sigma = 30\%$]{
    \raisebox{-.5\height}{\includegraphics[width=0.48\linewidth]{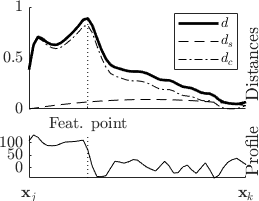}}
    }

    \subfloat[SLIC distance -- $\sigma = 60\%$]{
    \raisebox{-.5\height}{\includegraphics[width=0.48\linewidth]{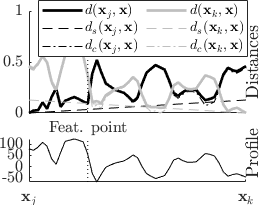}}
    }
    \subfloat[Robust saliency -- $\sigma = 60\%$]{
    \raisebox{-.5\height}{\includegraphics[width=0.48\linewidth]{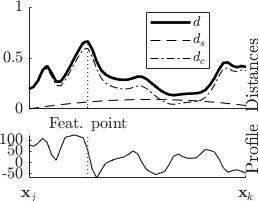}}
    }
    \caption{Comparison of the two salient point search strategies proposed. The top row shows the input images with different amounts of Gaussian noise ($\sigma$) as a fraction of the maximum image intensity. The edge of interest is highlighted in white. For the remaining rows, the left column shows, for each value of $\sigma$, two graphs: on the bottom, the intensity profile along that edge, and on the top, the SLIC distances (total distance, $d$, distance in space, $d_s$, and distance in colour, $d_c$) which should cross at the salient feature point. The right column shows the robust saliency measure, which is at maximum at the feature point.  Each row shows the intensity profile along the same edge $e$ between the points $\mathbf{x}_j$ and $\mathbf{x}_k$, when an increasing (from top to bottom) amount of noise is added.\label{fig:saliencyMeasure}} 
\end{figure}

\subsubsection{SLIC Distance-Based Feature Point Detection}
\label{sec:distancebased}

This approach searches for the feature point in a similar way as Simple Linear Iterative Clustering (SLIC) superpixels by \cite{Achanta2012} search for boundaries between pixel clusters. For each edge $e =(j,k) \in E$, we define that $\hat{ \mathbf{x} } \in \overline{\mathbf{x}_j \mathbf{x}_k}$ is on an image feature point if there is a distance measure $d$ for which $d(\mathbf{x}_j, \mathbf{x}^a)<d(\mathbf{x}^a, \mathbf{x}_k)$ for any $\mathbf{x}^a \in \overline{ \mathbf{x}_j  \hat{ \mathbf{x} } }$ and  $d(\mathbf{x}_j, \mathbf{x}^b)>d(\mathbf{x}^b,\mathbf{x}_k)$ for any $\mathbf{x}^b \in \overline{ \hat{ \mathbf{x} } \mathbf{x}_k}$. In other words, any point on the `left' of  $\hat{ \mathbf{x} }$ is closer to $\mathbf{x}_j$, and any point on its `right' is closer to $\mathbf{x}_k$.


We compute $d$ as a weighted sum of the distance in space and the distance in image intensity. Image data is defined as a discrete scalar function $I: \Omega \subset \mathbb{N}^d \longmapsto \mathbb{R}$, where $\Omega$ is the discrete grid where image pixels are defined. When required, image intensities $I(\mathbf{x})$ at non-grid positions $\mathbf{x} \notin \Omega$ are computed by linearly interpolating image values at neighbouring grid points. The distance in space $d_s(\mathbf{x}_j,\mathbf{x})$ is defined as the Euclidean distance on the graph edge:
\begin{equation}
	d_s(\mathbf{x}_j,\mathbf{x}) = || \mathbf{x}_j - \mathbf{x} ||
\end{equation}
and image intensity distance, $d_c$, is defined as the absolute difference in intensity from the edge nodes:
\begin{equation}
    d_c(\mathbf{x}_j,\mathbf{x}) = || I(\mathbf{x}_j) - I(\mathbf{x}) ||
\end{equation}

In order to ensure independence from edge length and local intensity range, $d_s$ is normalised by the edge length $N_s(\mathbf{x}_j, \mathbf{x}_k) = || \mathbf{x}_k - \mathbf{x}_j ||$, and $d_c$ is normalized by a parameter $N_c$. These two distance measures can be combined using a weighting factor $\lambda \in \mathbb{R}^+$:
\begin{equation}
	d(\mathbf{x}_j,\mathbf{x}) = \sqrt{ \left(\frac{d_c(\mathbf{x}_j,\mathbf{x})}{N_c}\right)^2 + \lambda\left(\frac{ d_s(\mathbf{x}_j,\mathbf{x})}{N_s(\mathbf{x}_j, \mathbf{x}_k)}\right)^2}
    \label{eq:distanceprima}
\end{equation}
The above equations can be analogously described for the other edge point, $\mathbf{x}_k$. As pointed out by \cite{Achanta2012}, variability of intensities throughout the image and between images make it difficult to calculate $N_c$. As a consequence, $N_c$ can be made constant to an arbitrary value and then the contribution of the intensity distance can be fixed for each image and application using the weighting factor $\lambda$ that controls the trade-off between the two distances. In practice, setting $N_c$ to the estimated image intensity range will set each term to be between $0$ and $1$ independently on the edge lengths and the image intensity along the edges. Lower values of $\lambda$ enforce image boundary adaptation while larger values of $\lambda$ push the salient point towards the centre of the edge. Using this formulation, the most salient image point on the graph edge can be found at the crossing of $d(\mathbf{x}_j,\mathbf{x})$ and $d(\mathbf{x}_k,\mathbf{x})$, as shown in the examples in Fig. \ref{fig:saliencyMeasure} (left column).

\subsubsection{Robust Feature Point Detection}
\label{sec:integrateddistance}

The intensity distance measure presented in Sec. \ref{sec:distancebased} is effective for pixel clusterization but can be very sensitive to noise (e.g. Fig. \ref{fig:saliencyMeasure}). For this reason, we introduce a robust alternative to find the most salient point along the graph edge, which is particularly useful for noisy images. Since regardless of the data dimensionality $d$ edges are always 1-dimensional, we define the parameter $t \in [0,1]$, which can be used to obtain the positions $\mathbf{x}_e(t)$ along the edge $e=(j,k)$ as follows: $\mathbf{x}_e(t)=\mathbf{x}_j + t*(\mathbf{x}_k-\mathbf{x}_j)$. Using this parameterization, we define the `sided integrated intensity' functions $v^j_e(t)$ and $v^k_e(t)$ for the edge $e$ as follows:
\begin{equation}
v_e^j(t) = \int_{t=0}^t f_e^j(t) I\left(\mathbf{x}_e(t)\right) dt
\label{eq:smoothedintensitydistance}
\end{equation}
and
\begin{equation}
v_e^k(t) = \int_{t}^{t=1} f_e^k(t) I\left(\mathbf{x}_e(t)\right) dt
\label{eq:smoothedintensitydistance2}
\end{equation}
where the sided weighting functions are $f_e^j(t) = 1/|| \mathbf{x}_e(t)-\mathbf{x}_j||$ and $f_e^k(t) = 1/|| \mathbf{x}_e(t)-\mathbf{x}_k||$. If we consider that $e$ crosses only one image contour, and that this crossing occurs at the edge point $\hat{\mathbf{x} }$, then $v_e^j(t)$ and $v_e^k(t)$  from equations (\ref{eq:smoothedintensitydistance}) and (\ref{eq:smoothedintensitydistance2}) represent, respectively, the average value of the image along $e$ at each side of the contour. As a result, the squared difference between the two will be maximum when the correct salient location is found, because it ensures the more distinct average values at each side. More formally, saliency along the edge $e$ can be described by the function $s_e(t)$:
\begin{equation}
s_e(t) =  \left(\frac{\int_{0}^t f_e^j(t) I\left(\mathbf{x}_e(t)\right) dt  - \int_{t}^1 f_e^k(t) I\left(\mathbf{x}_e(t)\right) dt}{N_c}\right)^2
\label{eq:saliencyrobust}
\end{equation}
On homogeneous regions, the desired behaviour is that the salient point is found at the center of the edge and with a low saliency value. To achieve this, we add a regularization term, $r(t)$, which is a convex parabola valued 0 at the nodes and 1 at the center, and therefore pushes the maximum towards the center of the edge:
\begin{equation}
r(t) = -4\left(t^2 - t\right)
\end{equation}
The complete edge saliency measure $m_e$ is
\begin{equation}
m_e(t) = s_e(t)+ \lambda r(t)
\end{equation}
The salient point can be found by maximizing $m_e$:
\begin{equation}
\hat{t} = \arg \max_t m_e(t)
\label{eq:maxsearch}
\end{equation}

And the salient point $\hat{\mathbf{x}_e} = \mathbf{x}_e(\hat{t})$.  In practice, for each edge, the image is sampled at regular intervals depending on the edge length in relation to the image resolution. As a result, $N_s$, which was previously introduced in Sec. \ref{sec:distancebased} to denote the edge length, is used to denote the number of samples per edge in practice. For the images used in this paper, a value of $N_s$ between 10 and 30 samples along each edge were used. As a result, equation (\ref{eq:maxsearch}) can be solved by exhaustive search with very little computational cost, O($N_s$). Note that when increasing dimensionality of the images from 2D to 3D or beyond, the only added computational cost would be that of higher order interpolation, but edges always remain linear structures in $n$-dimensions.
\subsection{Saliency Measure}
\label{sec:saliencyMeasure}

As described above, the graph adaptation process consists of a search for feature points in 1D. As a result, feature saliency is determined by the shape of the distance functions at the feature point. We define saliency in two different ways for the two feature point search strategies proposed:

\begin{itemize}
\item SLIC Distance-based feature point detection (Sec. \ref{sec:distancebased}): \hfill \\Salient features such as image contours will yield steeper crossings of the distance curves, because changes in intensity along the graph edge will be higher, as can be seen in the noise-less case in the top in Fig. \ref{fig:saliencyMeasure}. In this case, saliency is computed using the highest slope value of the two intensity distance curves at the point where they cross, i.e.:
\begin{equation}
 s_e(\hat{\mathbf{x}}) = \max\{ |\dot{d_c}(\mathbf{x}_j, \hat{\mathbf{x}_e})|, |\dot{d_c}(\mathbf{x}_k, \hat{\mathbf{x}_e})|\}/N_c
\end{equation}
where $\dot{d_c}$ is the derivative of $d_c$, indicating the slope of the distance at the crossing.

\item Robust feature point detection (Sec. \ref{sec:integrateddistance}): \hfill \\In this case, the saliency can be directly calculated using equation (\ref{eq:saliencyrobust}) as the value $s_e(\hat{t})$. This is the peak value in the curves on the right hand side column of Fig. \ref{fig:saliencyMeasure}.
\end{itemize}

Note that we only consider image intensity distance $d_c$ to compute saliency, because feature saliency should be independent of the location of the graph nodes relative to the salient point. 

\subsection{Iterative Graph Adaptation}
\label{sec:iterativeadaptation}

Once the salient points have been found with either of the distance-based or the robust feature point detection method, the nodes of the adapted graph are updated to the centroid of the salient points adjacent to each node. The process is described in Alg. \ref{alg:graphadaptation}. After adaptation, the average Euclidean distance between the graph nodes before and after adaptation is used as a measure of residual, $R$. This process can be carried out iteratively until $R$ reaches a maximum residual or after a specific number of iterations. 

\begin{algorithm}
\caption{Graph Adaptation}\label{alg:graphadaptation}
\begin{algorithmic}[1]
\State $\mathcal{G}=(E,V) \gets $ uniform triangulation
\State $R = \infty$ \Comment Initial residual
\Repeat
	\For{ $e \in E$}
		\State Find salient point $\hat{\mathbf{x}}_e$
	\EndFor
    \State $V_a = \emptyset$ \Comment{Adapted nodes} 
    \For{ $v \in V$}
    	\State $X_v = \{\hat{\mathbf{x}}_e : v \in e\}$
    	\State $\bar{v} = mean(X_v)$ \Comment{Compute centroid of adjacent salient points}
		\State $V_a \gets \bar{v}$
	\EndFor
    \State $R = mean(|| V-V_a||)$
    \State $V \gets V_a$ \Comment{Update node locations}
\Until{$R < R_{max}$}
\end{algorithmic}
\end{algorithm}

\subsection{Dual Graphs}
\label{sec:dualgraphs}

Edges of the adapted graph run across image contours but the desired behaviour of oversegmenting graphs is that edges run along image contours. For this, the oversegmenting graph is computed as the dual of the adapted graph.

Obtaining the dual graph ($\mathcal{G}_d=\{V_d,E_d\}$,  Fig. \ref{fig:example5}) from the adapted graph ($\mathcal{G}=\{V,E\}$, Fig. \ref{fig:example4}) is done as follows. The vertices of the dual graph $V_d$ are  the salient points found in $\mathcal{G}$. Edges $\{e_d\} \in E_d$ connect every two salient points $\hat{\mathbf{x}_{e_1}}$, $\hat{\mathbf{x}_{e_2}}$ which run along two edges $e_1, e_2 \in E$ that share a node in $V$. The edge saliency $s_{e_d}$ associated to $e_d$ can be computed as the product of the two saliency measures: $s_{e_d}= s(\hat{\mathbf{x}_{e_1}}) s(\hat{\mathbf{x}_{e_2}})$.

\subsection{Algorithmic Complexity}
\label{sec:complexity}

As there is one salient point calculation per graph edge, independently of the other edges, the algorithm is linear with the number of edges. As the number of edges $|E|$ is in turn linear with the number of nodes $|V|$, the complexity of the algorithm is $O(|V|)$. 

Since at each iteration, each edge is computed independently from the other edges, the algorithm would lend itself to an efficient parallel implementation, although this has been left for future work.

\section{Materials and Experiments}
\label{sec:experiments}

We carry out two types of experiments. The first type aims at assessing the ability of the oversegmenting graphs to adhere to image boundaries, which describe the structure of the image. The second typeaims at assessing the representation power of the adapted graphs for an image classification task.

\subsection{Experiments on Boundary Adherence and Oversegmentation}

We carry out experiments in synthetic images and on 200 images from the Berkeley Segmentation Data Set (BSDS300) by \cite{MartinAStatistics}. As is commonly done \citep{Achanta2012,Machairas2015}, we evaluate the performance of our method through the adherence to boundaries. We do this by approximating the boundary recall for the salient points as follows: a salient point $\hat{\mathbf{x}}$ is considered to lie on the boundary $\mathcal{C}$ (i.e. $\mathbf{ \hat{x} } \in \mathcal{C}$) if $d(\hat{\mathbf{x}},\mathcal{C})<d_{min}$, where $d$ is the distance to the boundary (computed through the morphological distance) and $d_{min}$ is a threshold set to two pixels, as in \cite{Achanta2012}. Using this convention, for each dual graph $\mathcal{G}_d = \{V_d, E_d\}$ we define the following sets:

\begin{equation}
\begin{array}{rcl}
\mathbf{T}(s_{min}) &=& \{\mathbf{ \hat{x} } \in V_d | s(\mathbf{ \hat{x} }) > s_{min} \}\\
\mathbf{P}(s_{min}) &=& \{\mathbf{ \hat{x} } \in\mathbf{T}(s_{min}) | dist(\mathbf{ \hat{x} }, \mathcal{C}) < d_{min} \} \\
\end{array}
\end{equation}
The boundary recall $R(s_{min})$ is defined as the fraction of salient points where saliency is greater than $s_{min}$ and that are at a distance smaller than $d_{min}$ from the image contour $\mathcal{C}$, and can be calculated using the cardinality of the above sets:
\begin{equation}
R(s_{min}) = \frac{| \mathbf{P}(s_{min}) |}{| \mathbf{T}(s_{min}) |}
\label{eq:recall}
\end{equation}
Note that the definition of recall in equation (\ref{eq:recall}) is different to the standard boundary recall definition for superpixels (i.e. fraction of boundary pixels that are coincident with superpixel edges), and therefore the two are not directly comparable.

\begin{figure}[htb!]
	\centering{}
	\subfloat[Diag]{
	\includegraphics[width=0.14\linewidth]{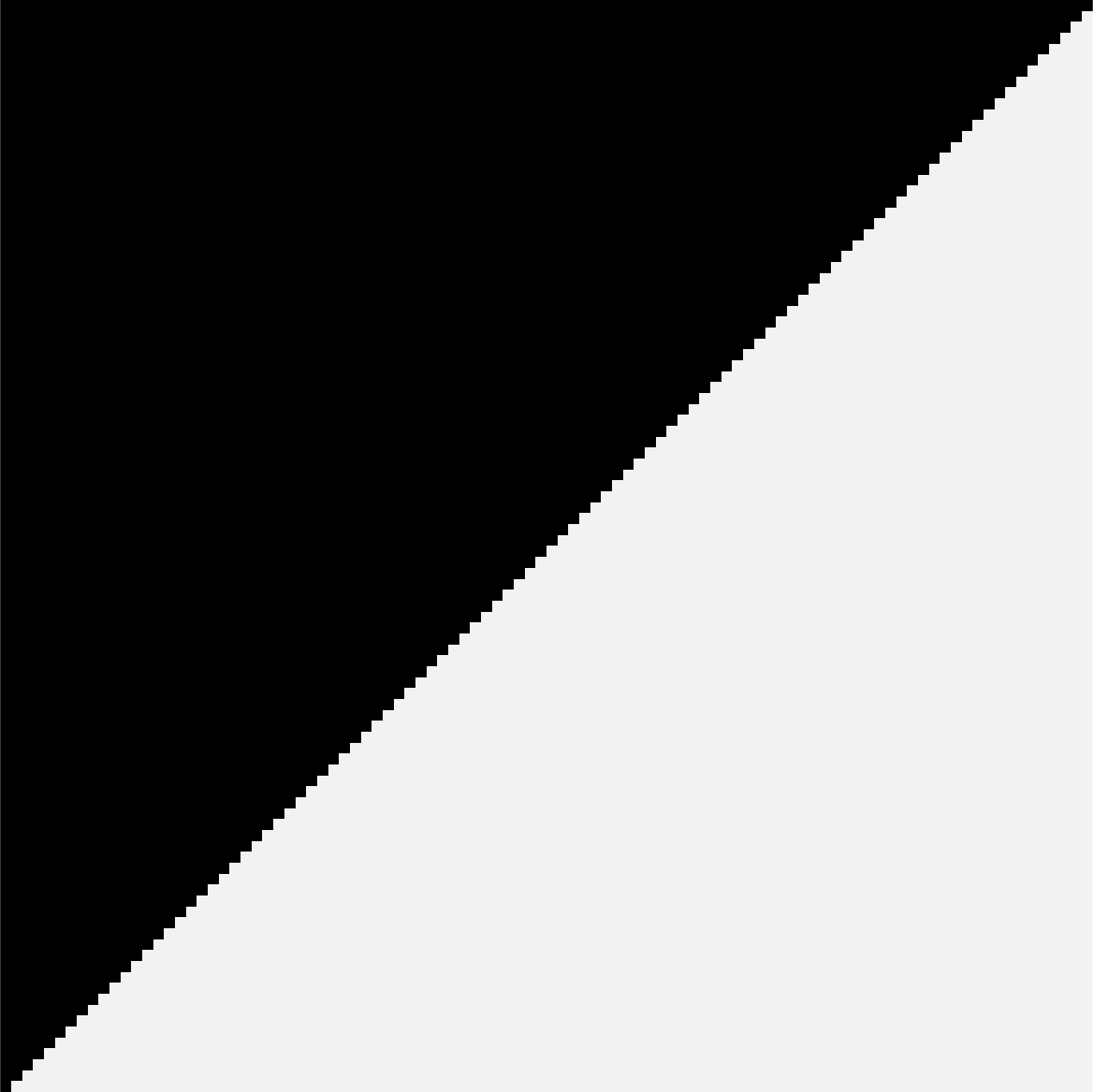}
    }
    \subfloat[Flat]{
	\includegraphics[width=0.14\linewidth]{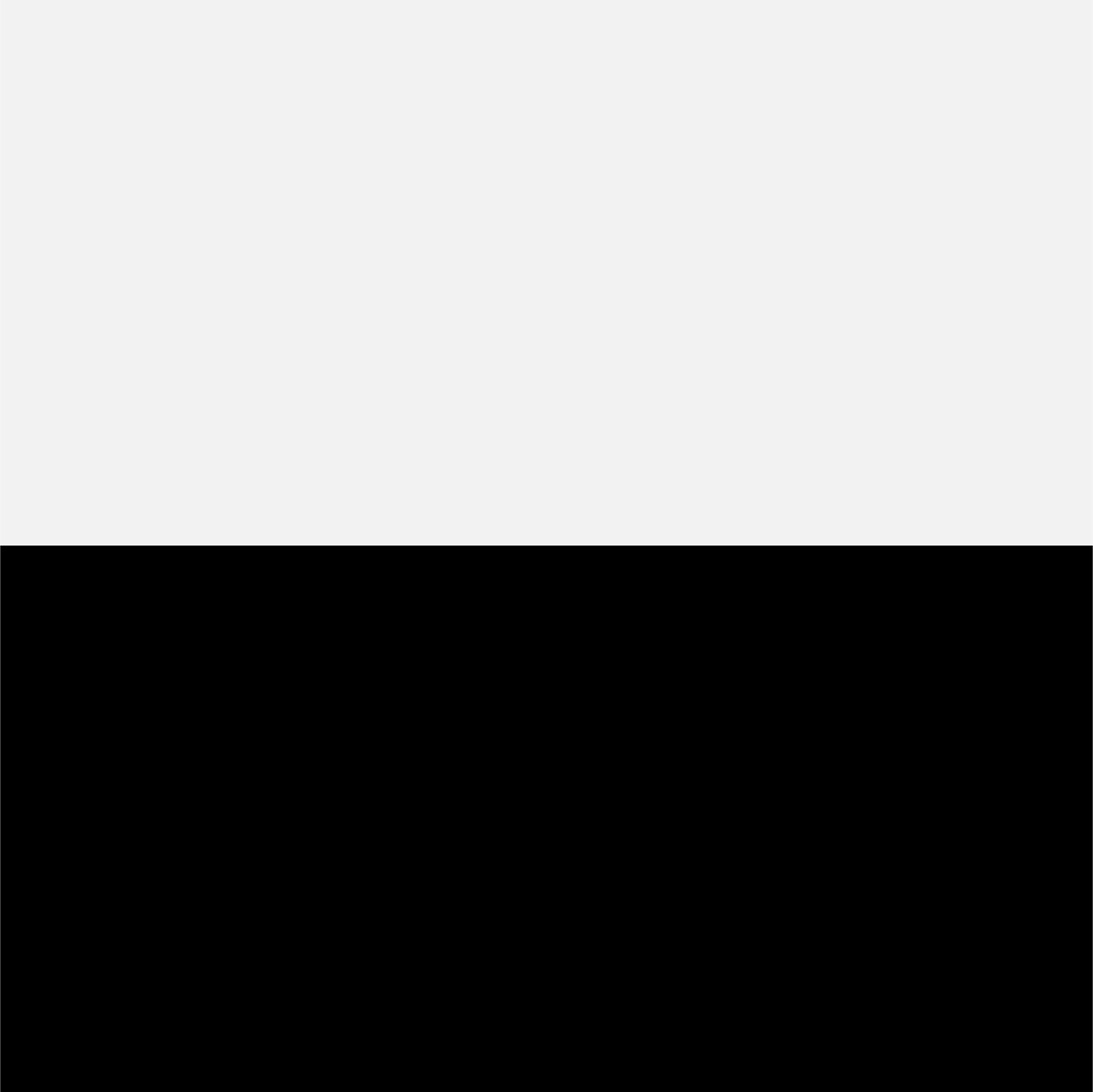}
    }
    \subfloat[Corner]{
	\includegraphics[width=0.14\linewidth]{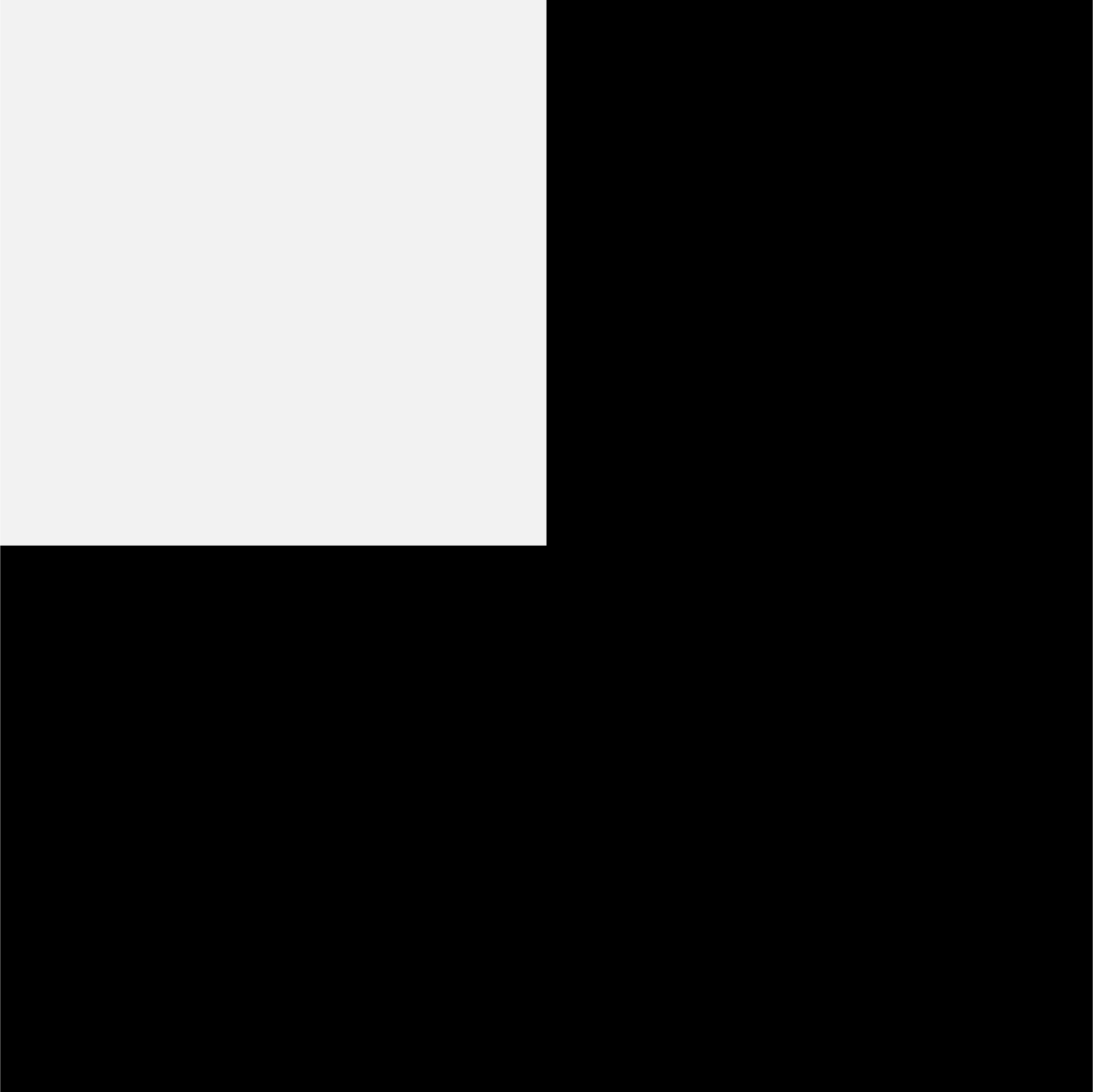}
    }
     \subfloat[Circle]{
	\includegraphics[width=0.14\linewidth]{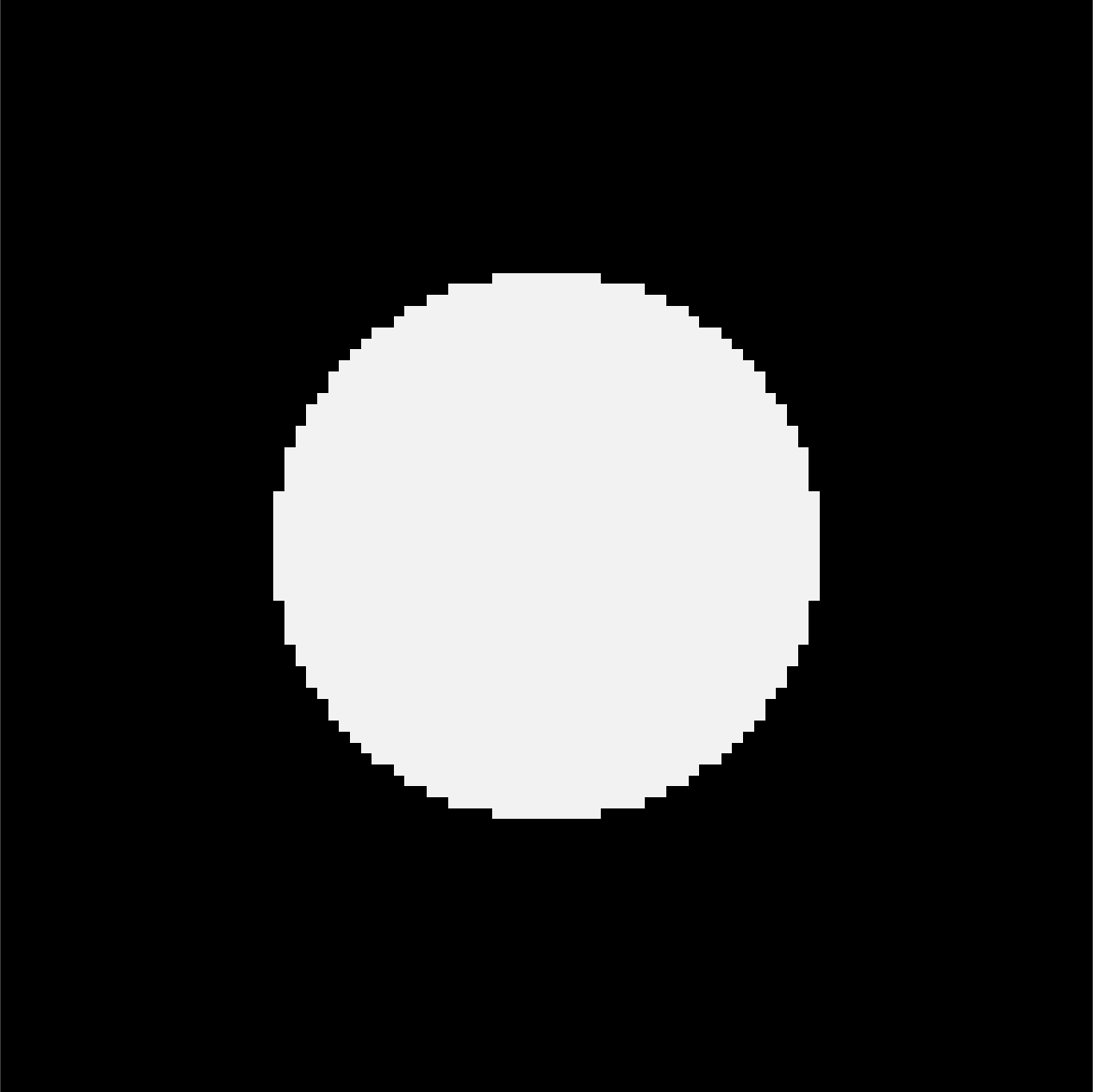}
    }
    \subfloat[Vertical]{
	\includegraphics[width=0.14\linewidth]{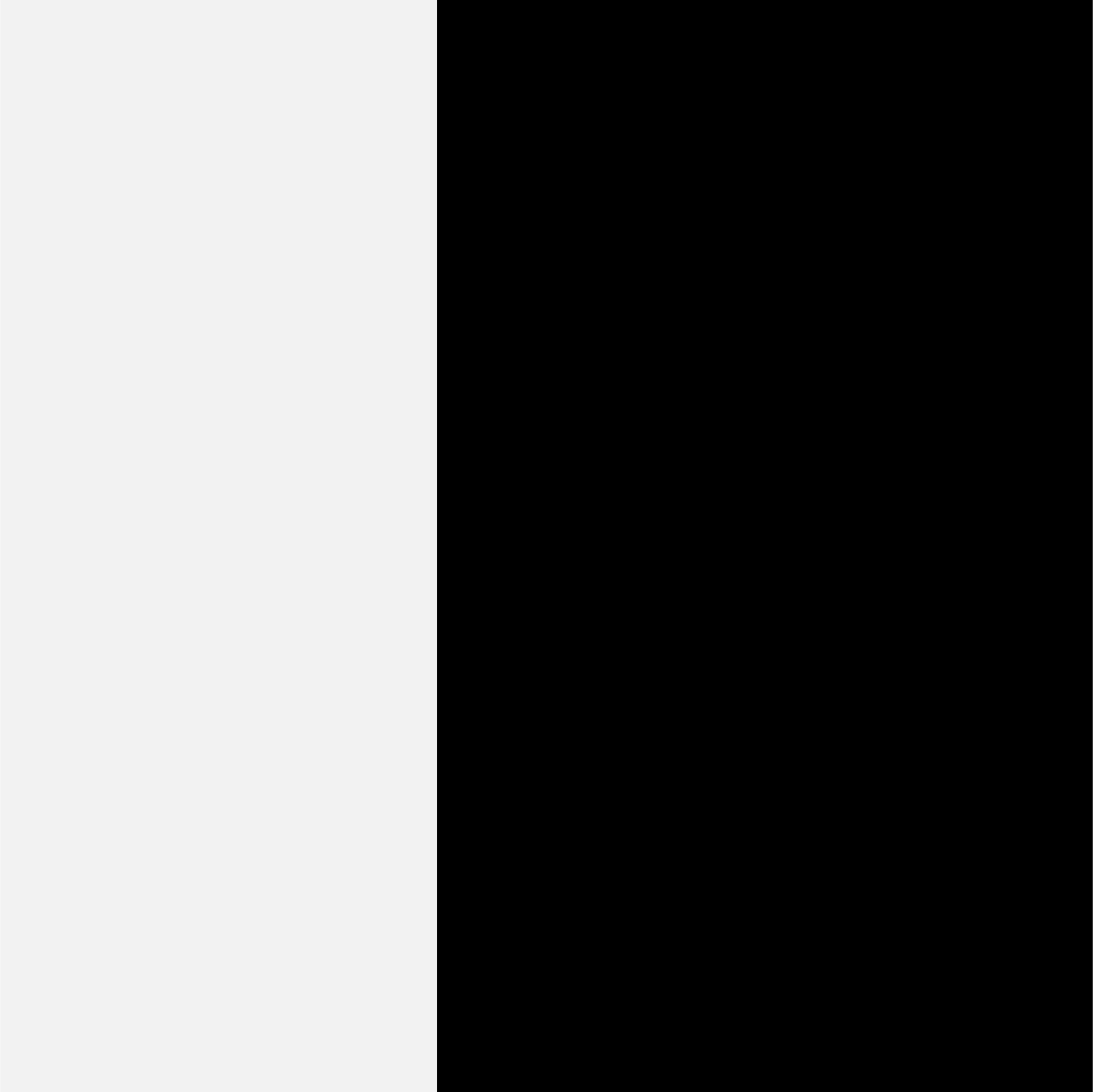}
    }
    \subfloat[Donut]{
	\includegraphics[width=0.14\linewidth]{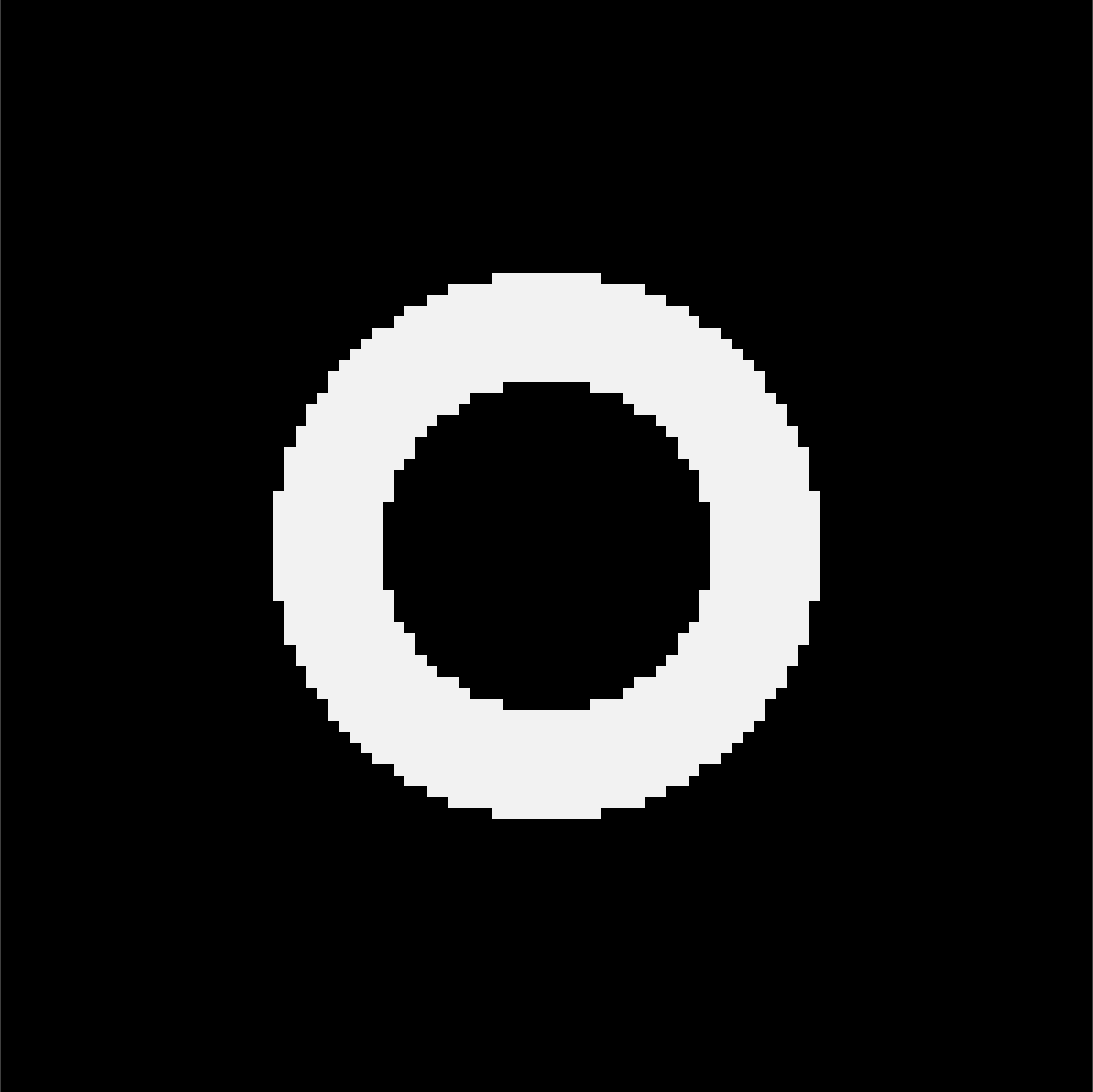}
    }
    \caption{Synthetic images with a range of shapes and boundaries. \label{fig:syntheticImages}} 
\end{figure}

For the experiments on synthetic data we generated six synthetic binary images, shown in Fig. \ref{fig:syntheticImages}, to investigate the performance of the proposed method using the two feature point detection strategies (SLIC distance based and robust), for image contours of different shapes and different noise levels. Additive centered Gaussian noise $\mathcal{N}(0,\sigma^2)$, with $\sigma$ between $0\%$ and $100\%$ the maximum image intensity value, was used.

In our experiments on natural images from BSDS300 we analysed the recall for a different number of nodes in the graph and different values of $s_{min}$. Finally, we also provide a qualitative comparison between the proposed  method and SLIC superpixel clustering \citep{Achanta2012}.

\subsection{Experiments on Graph-based Image Classification}

This second type of experiments aims to assess the ability of our proposed graphs to represent image content through a classification task. Recent work by \cite{fey2018splinecnn} proposed a novel geometric deep learning classification architecture, named SplineCNN, and used it to classify digits from the MNIST dataset using the usual split of 60K images for training and 10K images for testing. Graphs were derived from images by computing a 75 superpixel clusterization following the method by \cite{defferrard2016convolutional}. This produces graphs that have 75 nodes and 1260 edges, where the image intensity at the nodes position is used as node a feature (Fig. \ref{fig:example_of_graphs_for_dgl:sp}). We apply the SplineCNN to our proposed adapted graphs obtained from uniform grids with $N=49,64,81,100,121$ nodes and $E=120,161,208,261,320$ edges respectively (Fig \ref{fig:example_of_graphs_for_dgl}). We test three node features with our graphs: using image intensity (as for the superpixels), using saliency, and using both. The architecture was adapted from the implementations publicly available from \cite{Fey/Lenssen/2019}.

\section{Results}
\label{sec:results}

\subsection{Boundary Adherence}

\subsubsection{Synthetic Images} 

\begin{figure}[htb!]
	\centering
    \subfloat[SLIC distance]{ 
	\includegraphics[width=\linewidth]{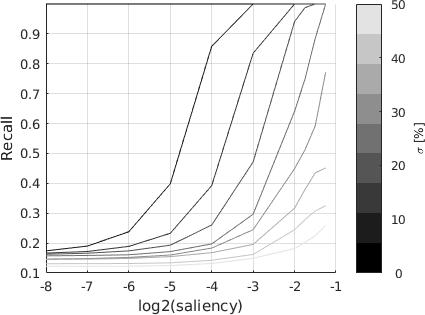}
    }
    
    \subfloat[Robust saliency]{
	\includegraphics[width=\linewidth]{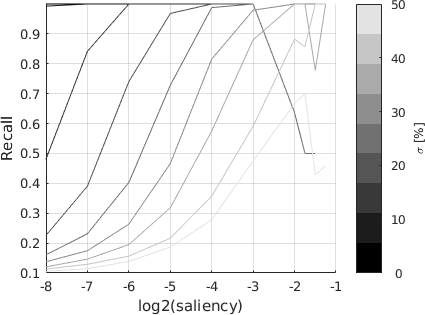}
    }
   
    \caption{Quantitative results on the synthetic data. The curves show the boundary recall for salient points beyond a certain saliency (in logarithmic units), gray level indicates the amount $\sigma$ of centred Gaussian noise, as a percentage of the maximum image intensity value. It can be observed that the curves corresponding to the robust saliency measure are shifted to the left, indicating that for the same amount of Gaussian noise, between a 50\% and a 70\% higher recall is achieved. \label{fig:resultstestimages}} 
\end{figure}

Figure \ref{fig:resultstestimages} shows the boundary recall curves for the two proposed methods (SLIC distance on top and robust saliency on the bottom) used to compute the oversegmenting graphs from the synthetic images from Fig. \ref{fig:syntheticImages}. The curves were obtained as an average over different values of the number of graph nodes ($K=64,...,400$). The gray scale indicates the amount of centered Gaussian noise added, ranging from $\sigma=0$ in black to $\sigma = 100\%$ the maximum image intensity in light gray. The horizontal axis indicates the saliency threshold (in logarithmic units) beyond which salient points were considered for the boundary recall. 

It can be observed that the curves using the SLIC distance follow a similar trend to the curves using the robust saliency but, are shifted towards lower saliency values. In other words, using the robust saliency measure, the boundary recall obtained for the same $\sigma$ is between $50\%$ and $70\%$ greater.

\begin{figure}[htb!]
	\centering{}
    \includegraphics[width=\linewidth]{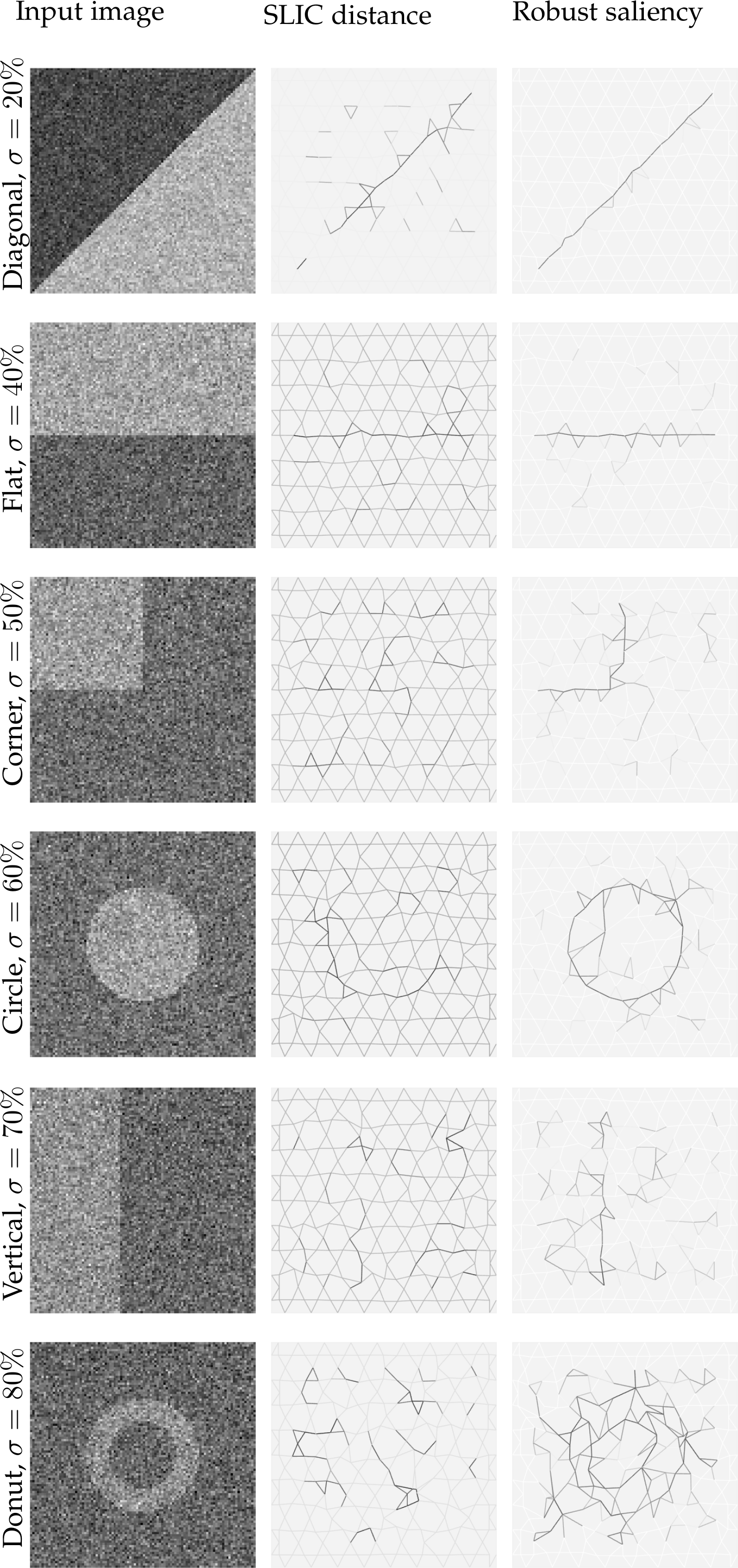}
    \caption{Example oversegmenting graphs on the synthetic images. The left column shows the input images to which a centered Gaussian noise of $\sigma = 20\%,...,80\%$ the maximum image intensity has been added. The centre column and the right hand side column show the oversegmenting graphs achieved using the SLIC distance and the robust distance, respectively, and colour-coded by $log_{2} (s)$.  \label{fig:resultstestimagesq}} 
\end{figure}

These results are consistent with the qualitative results shown in Fig. \ref{fig:resultstestimagesq}. Six representative examples of oversegmenting graphs obtained using the two proposed methods on input images with varying geometry and amount of noise are shown. All these examples were produced using $\lambda = 0.4$ and $K=100$ graph nodes. The resulting graphs are colour-coded by saliency (white indicates low saliency and black high saliency).

It can be observed that while for smaller amounts of noise (first row of Fig. \ref{fig:resultstestimagesq}, $\sigma=20\%$) the results with the two methods are fairly similar, the results using the robust saliency measure outperform those obtained with the SLIC distance, both in terms of saliency encoding (a more compact saliency can be observed along the true image edges) and in terms of adaptation ability. For example this is most obious in the `donut' figure at the bottom, where the results using the SLIC distance not only are unable to accurately highlight the true image shape but also the adapted edges do not follow that shape.

\subsubsection{BSDS300 Images}

\begin{figure}[htb!]
	\centering
    \subfloat[SLIC distance]{
	\includegraphics[width=\linewidth]{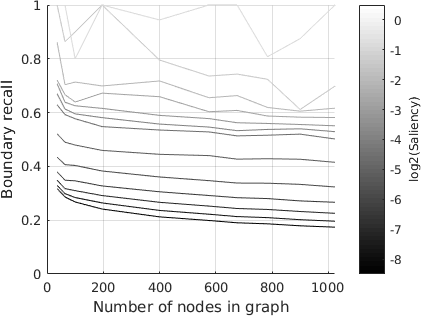}
    }

    \subfloat[Robust saliency measure]{
	\includegraphics[width=\linewidth]{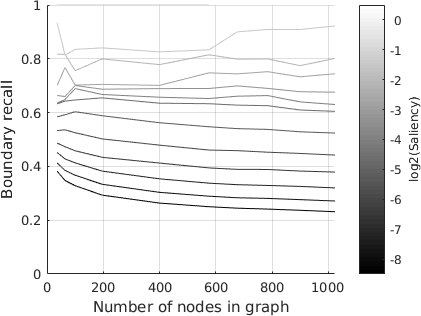}
    }
   
    \caption{Quantitative results on the BSD300 dataset. The curves show the boundary recall achieved for different number of nodes in the graph. The colour code indicates the minimum saliency value of the salient points considered for the recall computation, using a logarithmic scale.  \label{fig:resultsBSDS300}} 
\end{figure}

Figure \ref{fig:resultsBSDS300} shows the boundary recall curves for the two proposed methods (SLIC distance on the top, robust saliency on the bottom) in natural images from the BSDS300 database. The horizontal axis shows the number of graph nodes used, and the colour code indicates the saliency threshold beyond which salient points are considered to compute the boundary recall. 

For the two methods, as expected, higher saliency thresholds yield a higher recall for all number of nodes in the graph. Consistently with the results on synthetic images, the robust saliency measure yields higher recall by approximately $10\%$. 

\begin{figure*}[!b]
	\centering
    %
    %
    \subfloat[]{
	\includegraphics[height=0.22\linewidth]{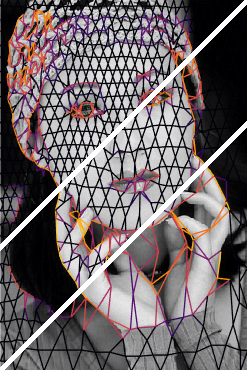}
    }
    \subfloat[]{
	\includegraphics[height=0.22\linewidth]{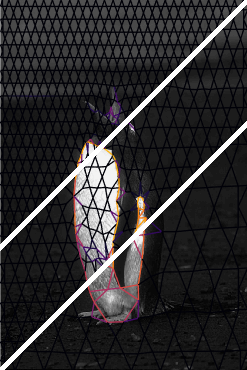}
    }
    \subfloat[]{
	\includegraphics[height=0.22\linewidth]{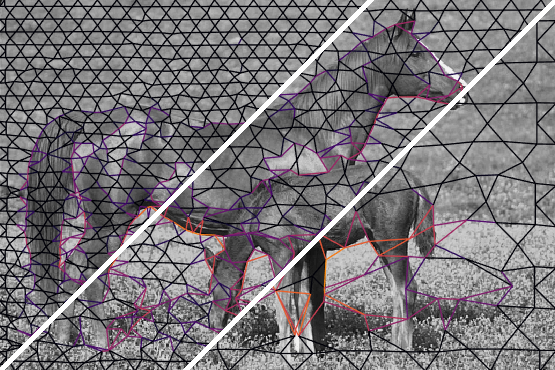}
    }
    \subfloat[]{
	\includegraphics[height=0.22\linewidth]{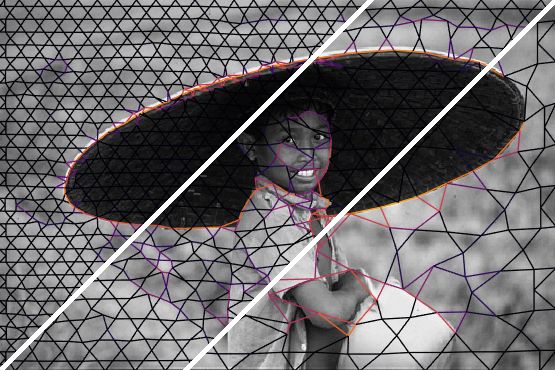}
    }
    \subfloat{
	\includegraphics[height=0.22\linewidth]{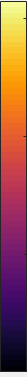}
    }
    
    \caption{Example results on the BSD300 using the robust salient point detection. The white lines are used to show the results at three different resolutions, for $K=\{100, 400, 900\}$ nodes. Saliency $s$ is color-coded using $log_2(s)$ and a color range from $-15$ (black) to $0$ (yellow).\label{fig:resultsExamples}} 
\end{figure*}

Figure \ref{fig:resultsExamples} shows the resulting oversegmenting graphs on four images from the BSDS300 database using the robust saliency measure. Each image shows the graph at three different resolutions (100, 400 and 900 grid nodes). The graph edges are colour-coded by the saliency, in logarithmic scale. It can be observed that the high saliency edges are consistent with apparent image contours, and that particularly at fine scales the graph follows image contours.

\subsubsection{Relation to Superpixel Clustering}

Superpixel clustering methods produce an oversegmentation of the input image by grouping together (clustering) pixels with similar intensities that are connected. Because of our choice of initial graph as a uniform triangular topology, the dual oversegmenting graph yields a partition of the space into hexagons and triangles (e.g. Figs. \ref{fig:example} and \ref{fig:resultsExamples}) and therefore the compactness of these regions is guaranteed by  definition. In the case of superpixels, connectivity is normally only enforced \citep{Achanta2012,Machairas2015}; (although in practice can be ensured given a sufficiently high regularization). While superpixels aim at creating groups of pixels that snap to image boundaries, our proposed method aims at snapping points at image boundaries and connecting those points, hence in practice snapping edges to image boundaries. This difference is of importance for two reasons: first, our proposed method does a 1D search regardless of the image dimensionality, while superpixel methods do an $N$D search; and second, superpixel methods label pixels in the image (without any explicit measure of how different two neighbouring regions are) while our method labels edges in a graph, and actually provides a saliency values for these edges which measures how different two image regions are at each side of the graph edge. 

\begin{figure*}[!b]
   
    \subfloat[SLIC superpixels \cite{Achanta2012}]{
	\includegraphics[width=0.31\linewidth]{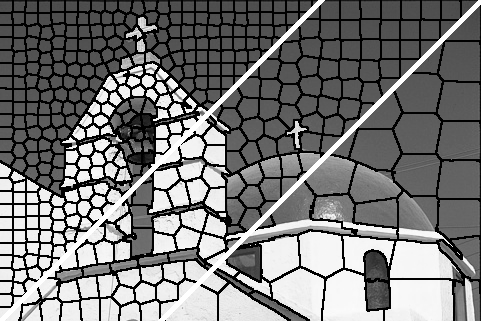}
    }
    \subfloat[Distance-based]{
	\includegraphics[width=0.31\linewidth]{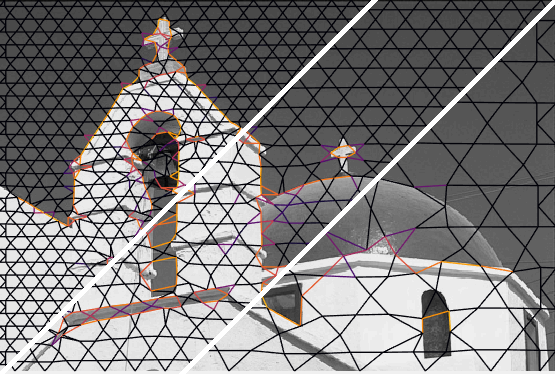}
    }
    \subfloat[Robust]{
	\includegraphics[width=0.31\linewidth]{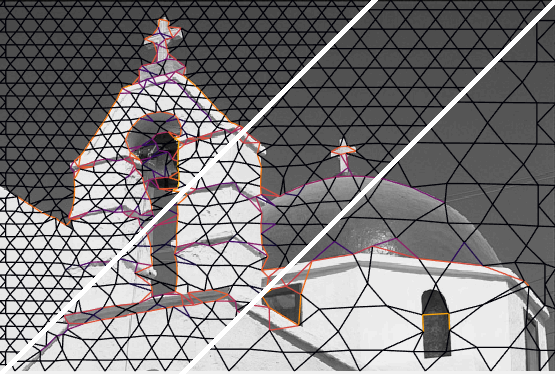}
    }
    \subfloat{
	\includegraphics[width=0.015\linewidth]{figures/colorbar}
    }
    
    \subfloat[SLIC superpixels \citep{Achanta2012}]{
	\includegraphics[width=0.31\linewidth]{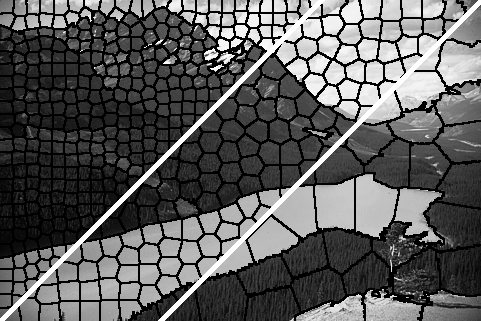}
    }
    \subfloat[Distance-based]{
	\includegraphics[width=0.31\linewidth]{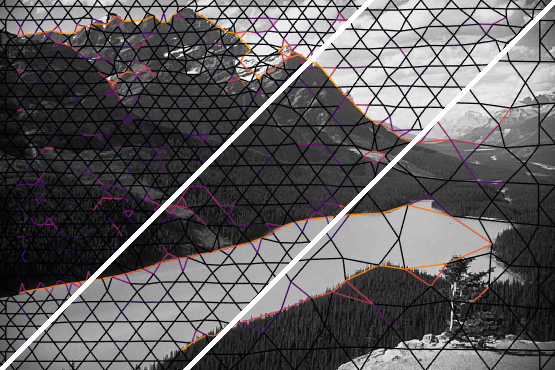}
    }
    \subfloat[Robust]{
	\includegraphics[width=0.31\linewidth]{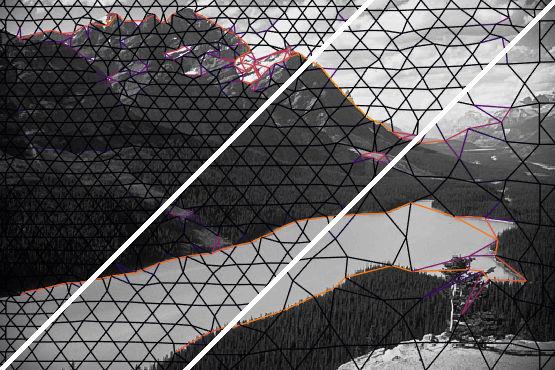}
    }
    \subfloat{
	\includegraphics[width=0.015\linewidth]{figures/colorbar}
    }
   
    \caption{SLIC Superpixel clustering by \cite{Achanta2012} vs proposed approach using the Distance-based and the Robust point saliency detection methods. Images are shown at three resolutions, with approximately $K=64, 400, 1000$ (number of seeds in SLIC and number of graph nodes in our proposed methods).\label{fig:resultsBSDS300visual}} 
\end{figure*}

Although superpixels and oversegmenting graphs are different in nature, they can yield visually similar results. Figure \ref{fig:resultsBSDS300visual} shows a qualitative comparison of SLIC superpixels by \cite{Achanta2012} with our oversegmenting graphs computed at different resolutions for the 3 example images from the BSDS300 database. It can be observed that our proposed method appears to converge to SLIC superpixels as the number of grid nodes increases, but in addition also provides a saliency measure for graph edges that are consistent with image contours.

\subsection{Graph-based Image Classification using Geometric Deep Learning}

\begin{figure*}[htb!]
	\centering
    \subfloat[SP $N=75$\label{fig:example_of_graphs_for_dgl:sp}]{
	\includegraphics[width=0.15\linewidth]{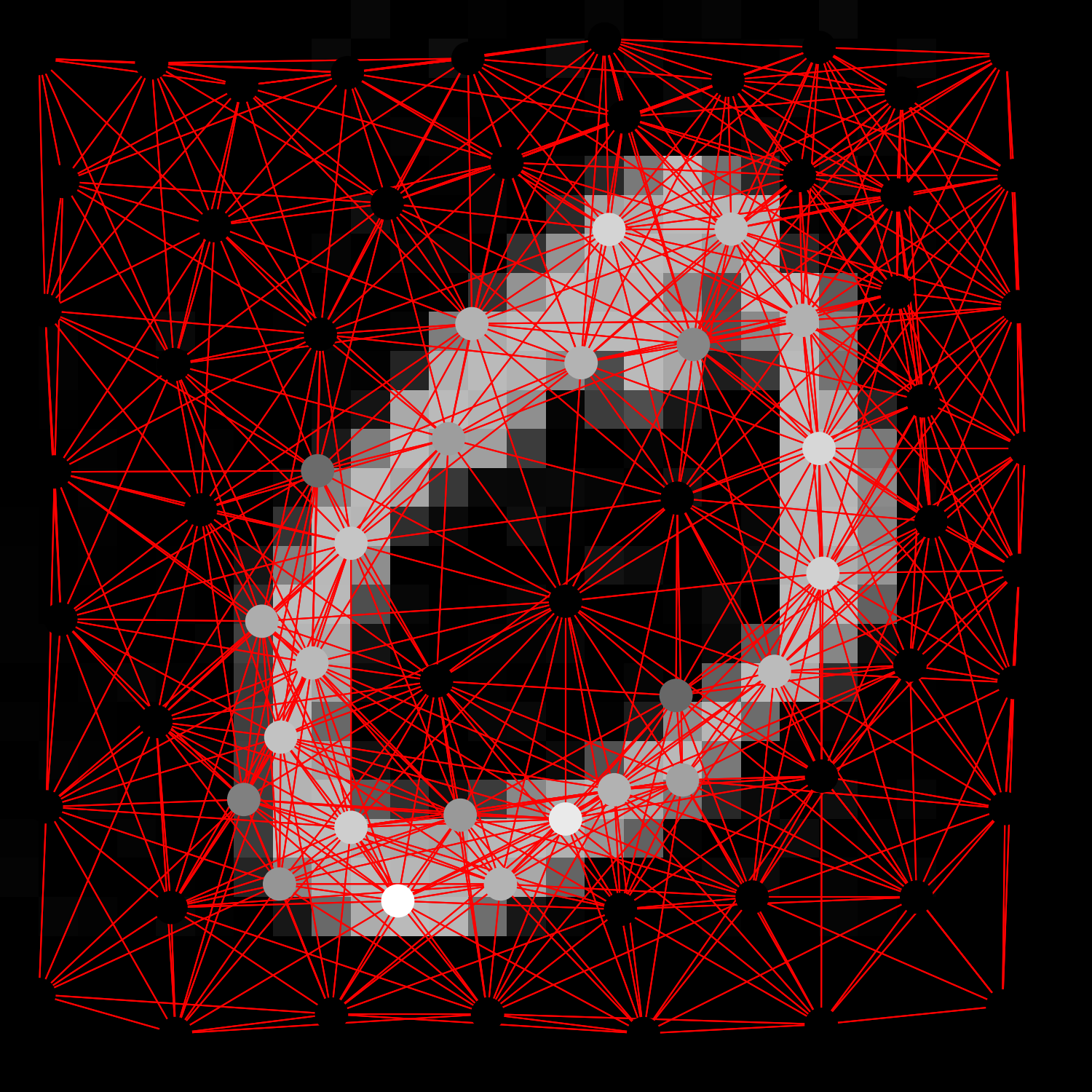}
    }
    \subfloat{
    \begin{minipage}[c]{0.01\linewidth}
    \rotatebox[origin=l]{90}{\ \ \ \ \ \ \ \ \ \ Intensity}
    \end{minipage}
    }
    \subfloat[AG $N=49$ \label{fig:example_of_graphs_for_dgl:49:0}]{
	\includegraphics[width=0.15\linewidth]{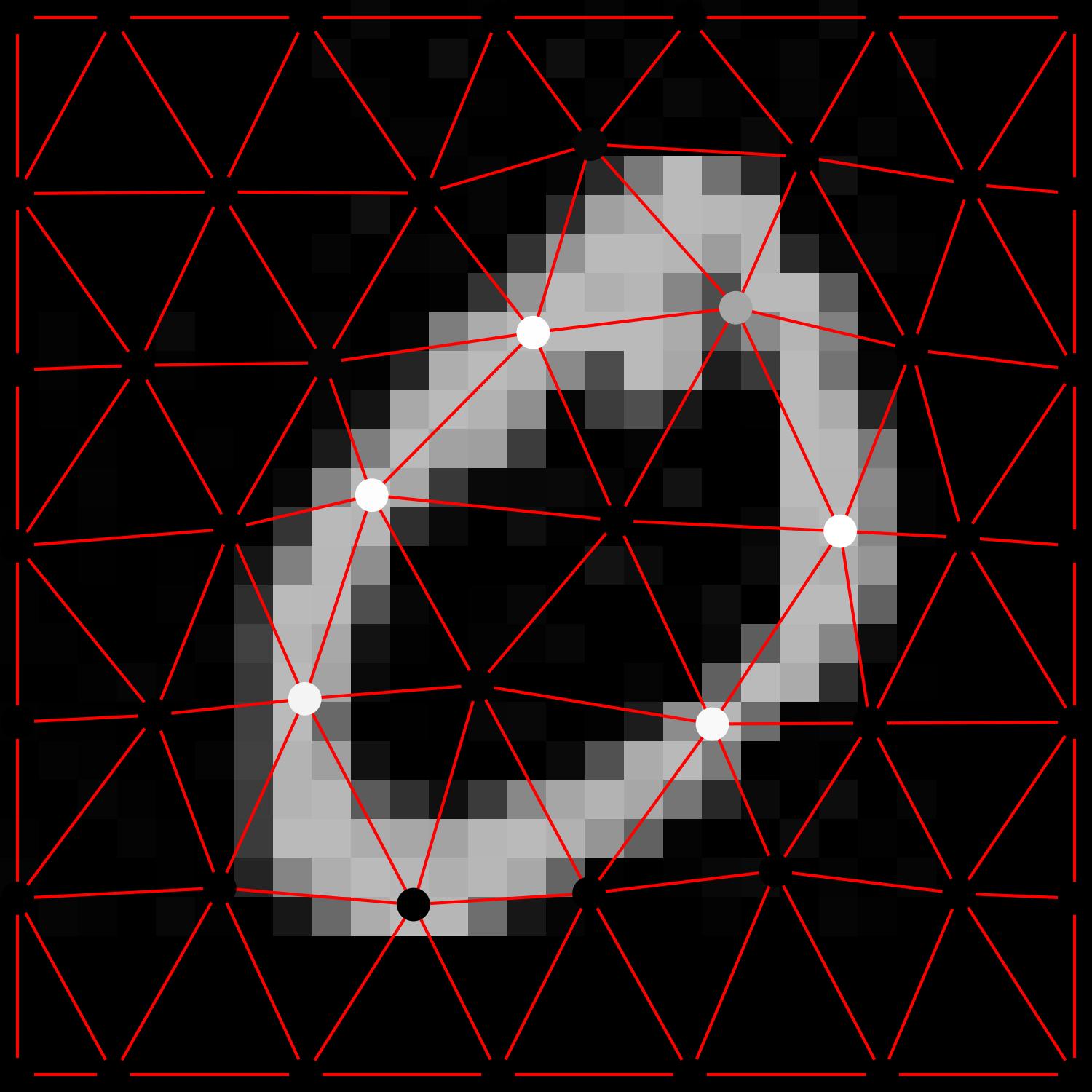}
    }
    \subfloat[AG $N=64$ \label{fig:example_of_graphs_for_dgl:64:0}]{
	\includegraphics[width=0.15\linewidth]{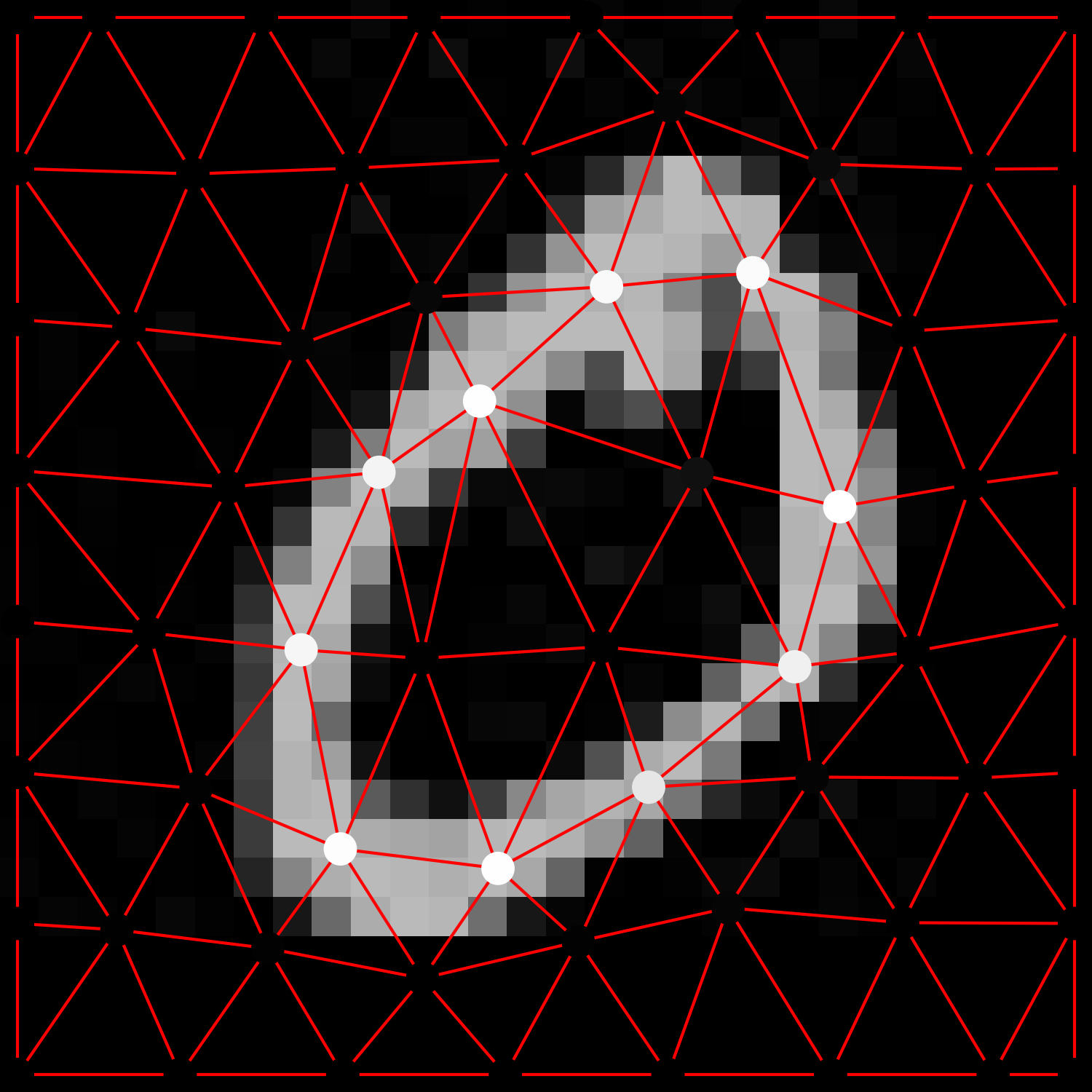}
    }
    \subfloat[AG $N=81$\label{fig:example_of_graphs_for_dgl:81:0}]{
	\includegraphics[width=0.15\linewidth]{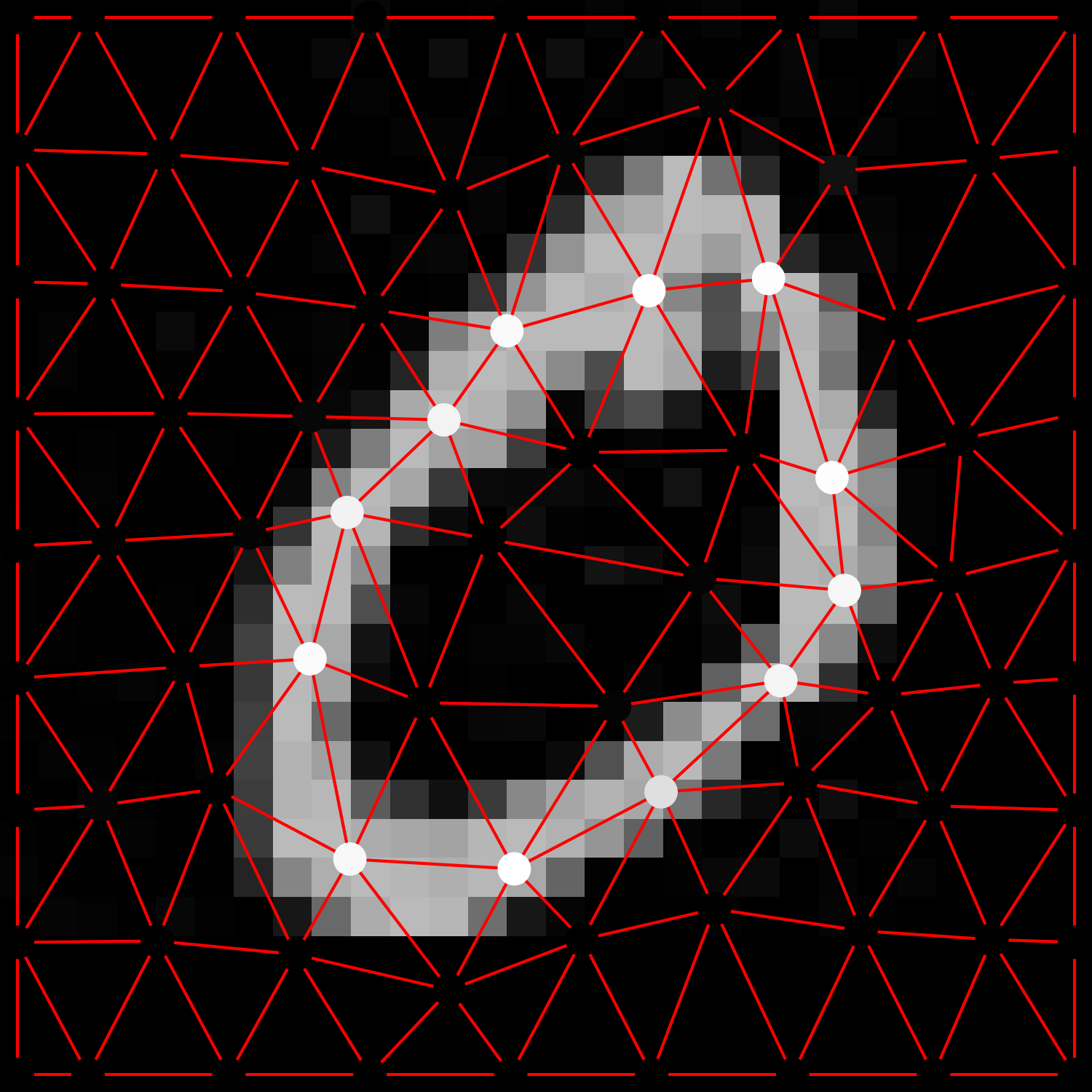}
    }
    \subfloat[AG $N=100$\label{fig:example_of_graphs_for_dgl:100:0}]{
	\includegraphics[width=0.15\linewidth]{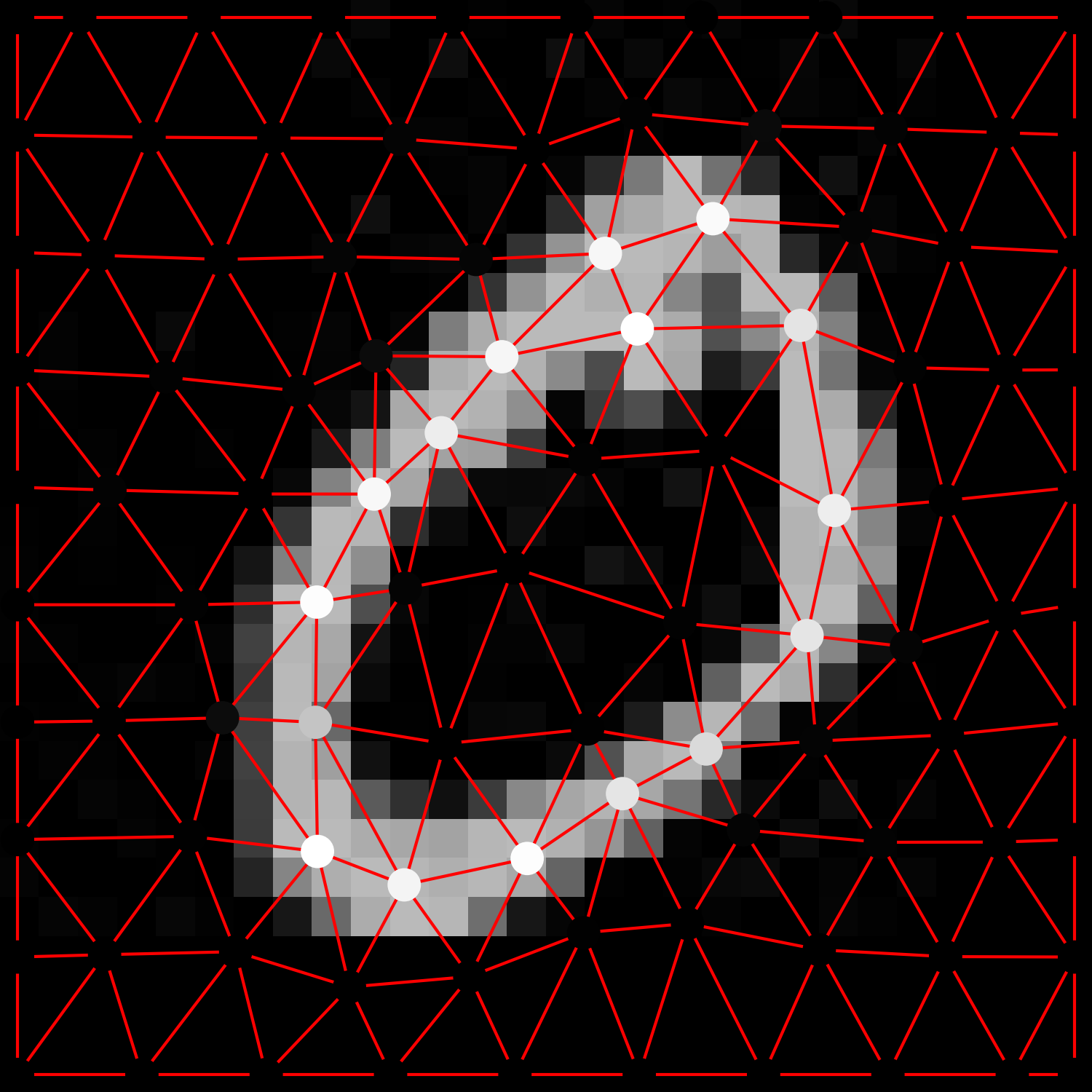}
    }
    \subfloat[AG $N=121$\label{fig:example_of_graphs_for_dgl:121:0}]{
	\includegraphics[width=0.15\linewidth]{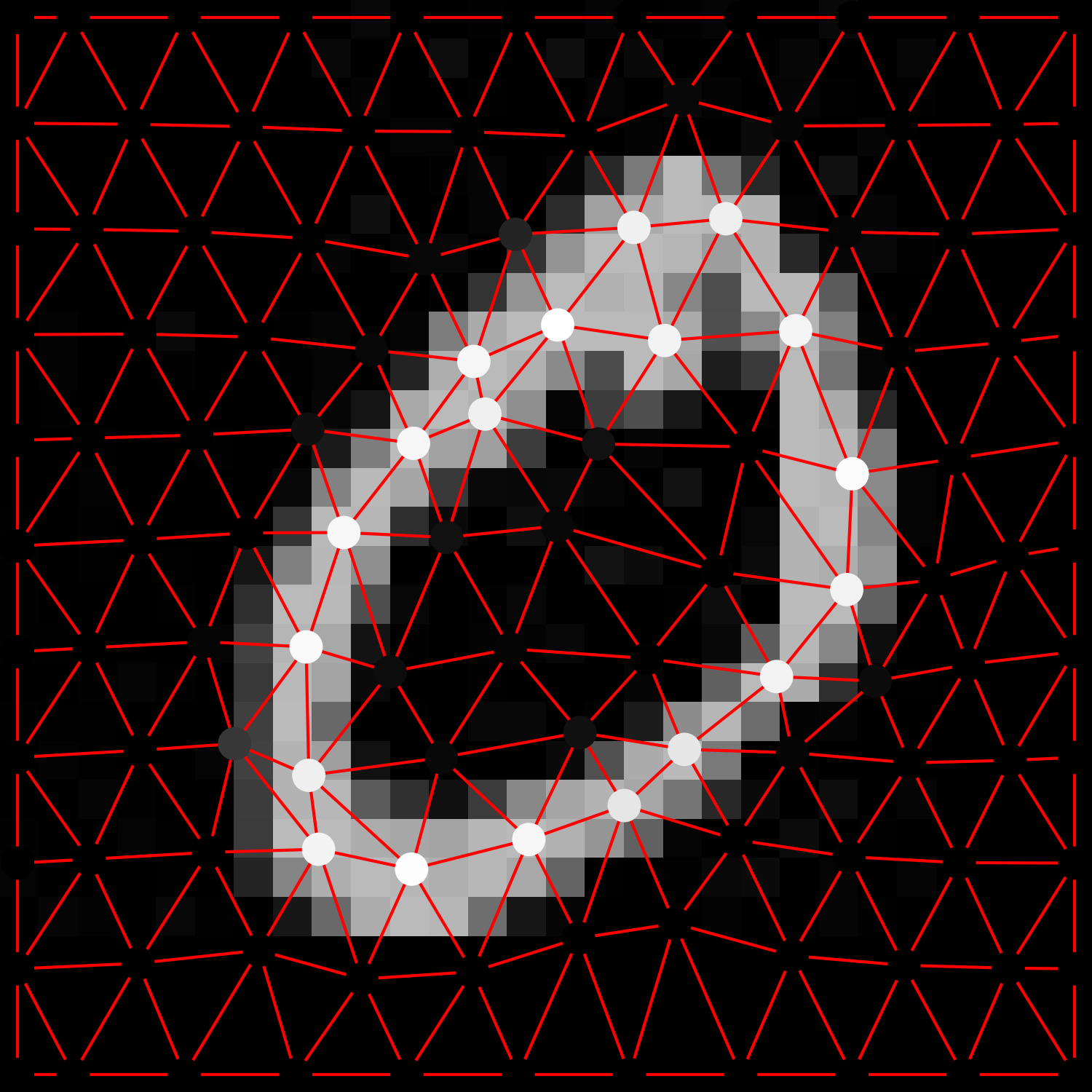}
    }    
    
    \vspace{-1cm}
    
    \hspace{0.15\linewidth}
    \subfloat{
    \begin{minipage}[c]{0.01\linewidth}
    \rotatebox[origin=l]{90}{\ \ \ \ \ \ \ \ \ \ \ Saliency}
    \end{minipage}
    }
    \subfloat[AG $N=49$ \label{fig:example_of_graphs_for_dgl:49:1}]{
	\includegraphics[width=0.15\linewidth]{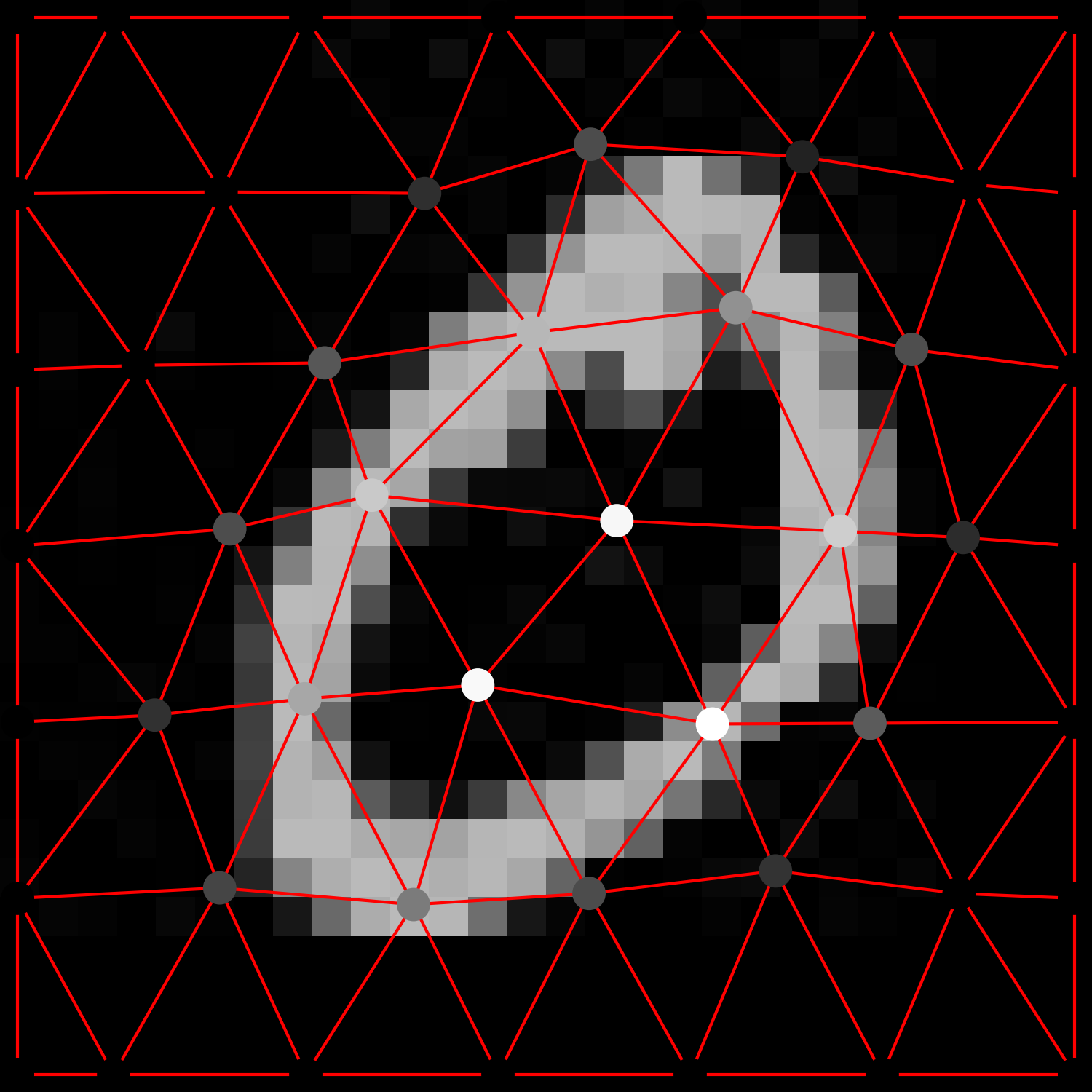}
    }
    \subfloat[AG $N=64$ \label{fig:example_of_graphs_for_dgl:64:1}]{
	\includegraphics[width=0.15\linewidth]{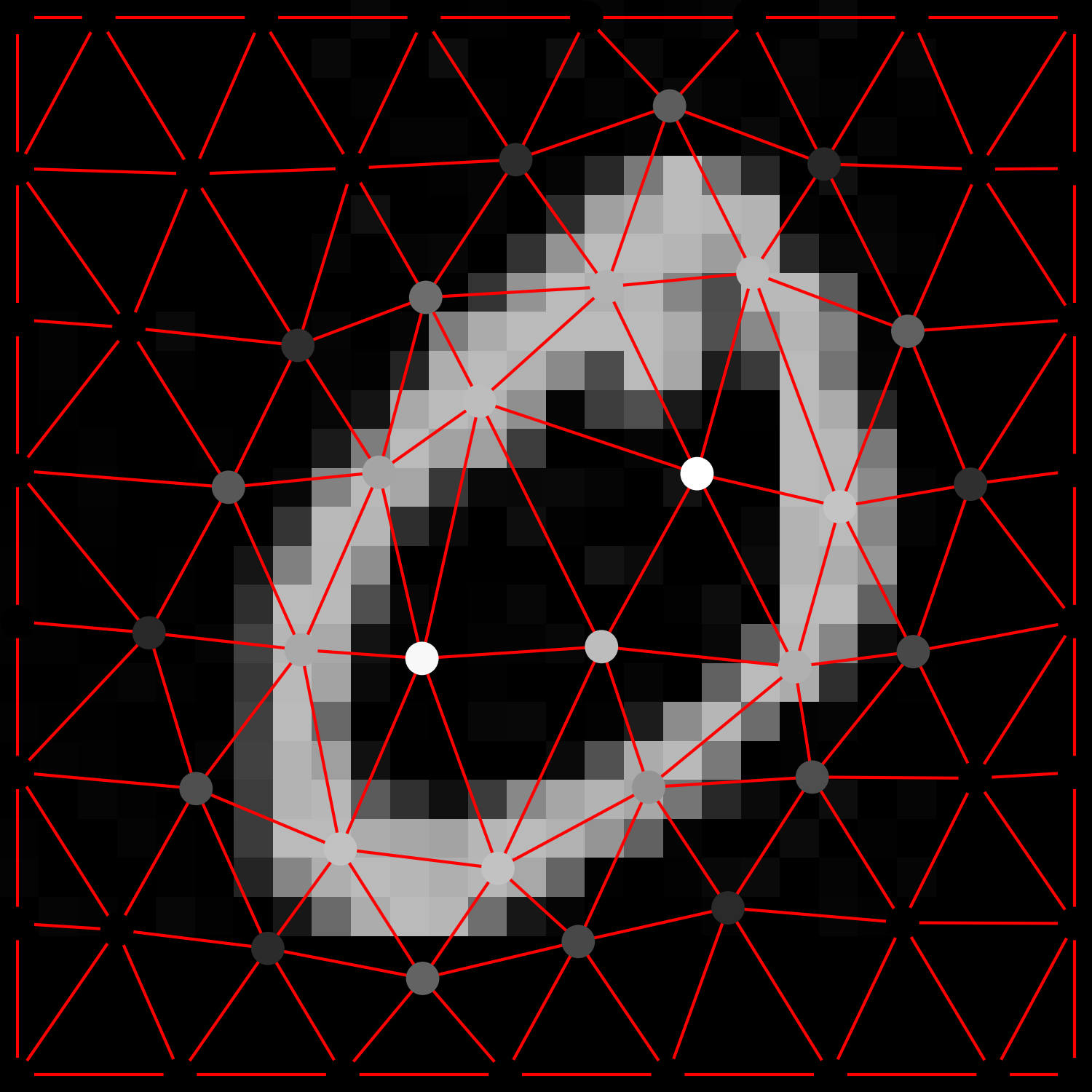}
    }
    \subfloat[AG $N=81$\label{fig:example_of_graphs_for_dgl:81:1}]{
	\includegraphics[width=0.15\linewidth]{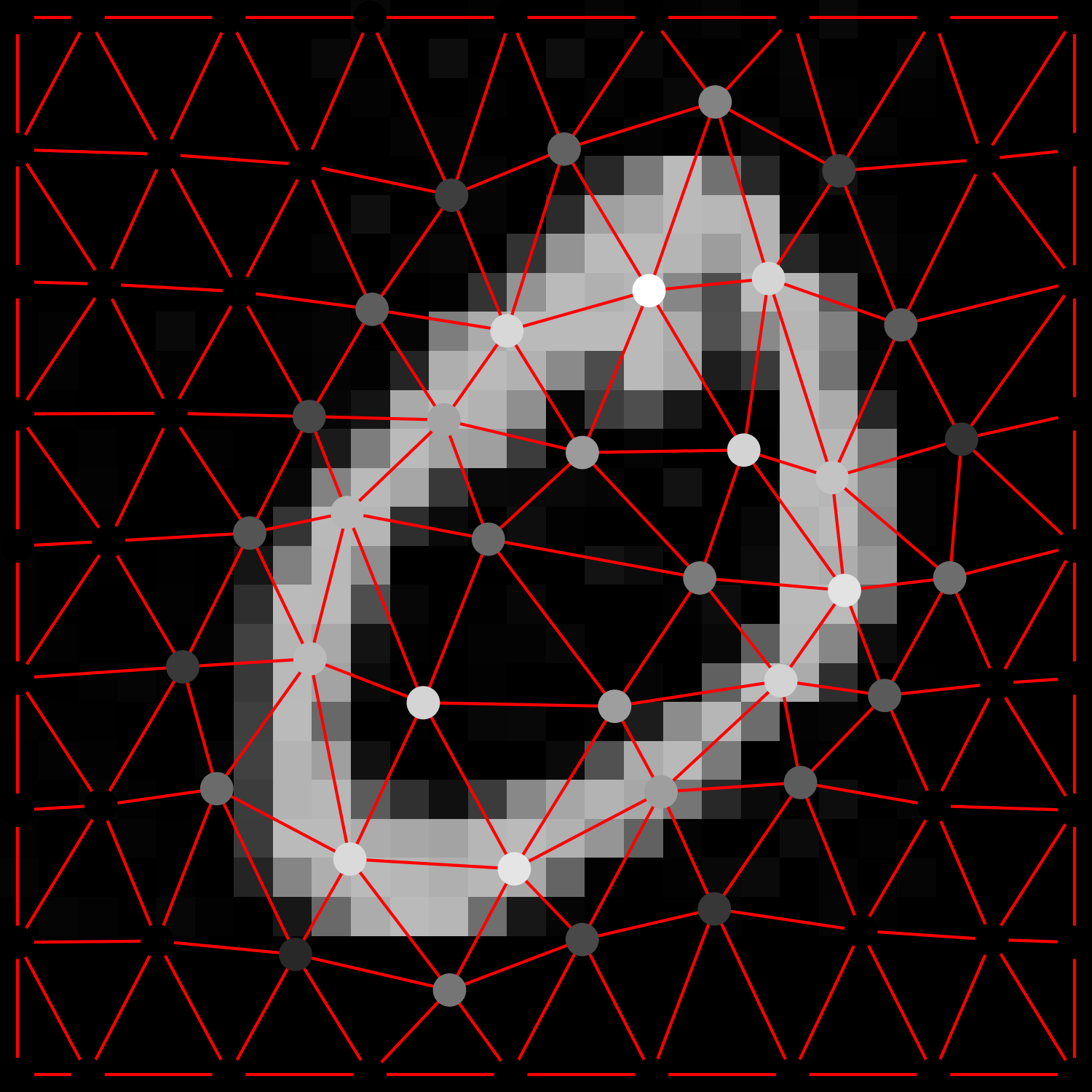}
    }
    \subfloat[AG $N=100$\label{fig:example_of_graphs_for_dgl:100:1}]{
	\includegraphics[width=0.15\linewidth]{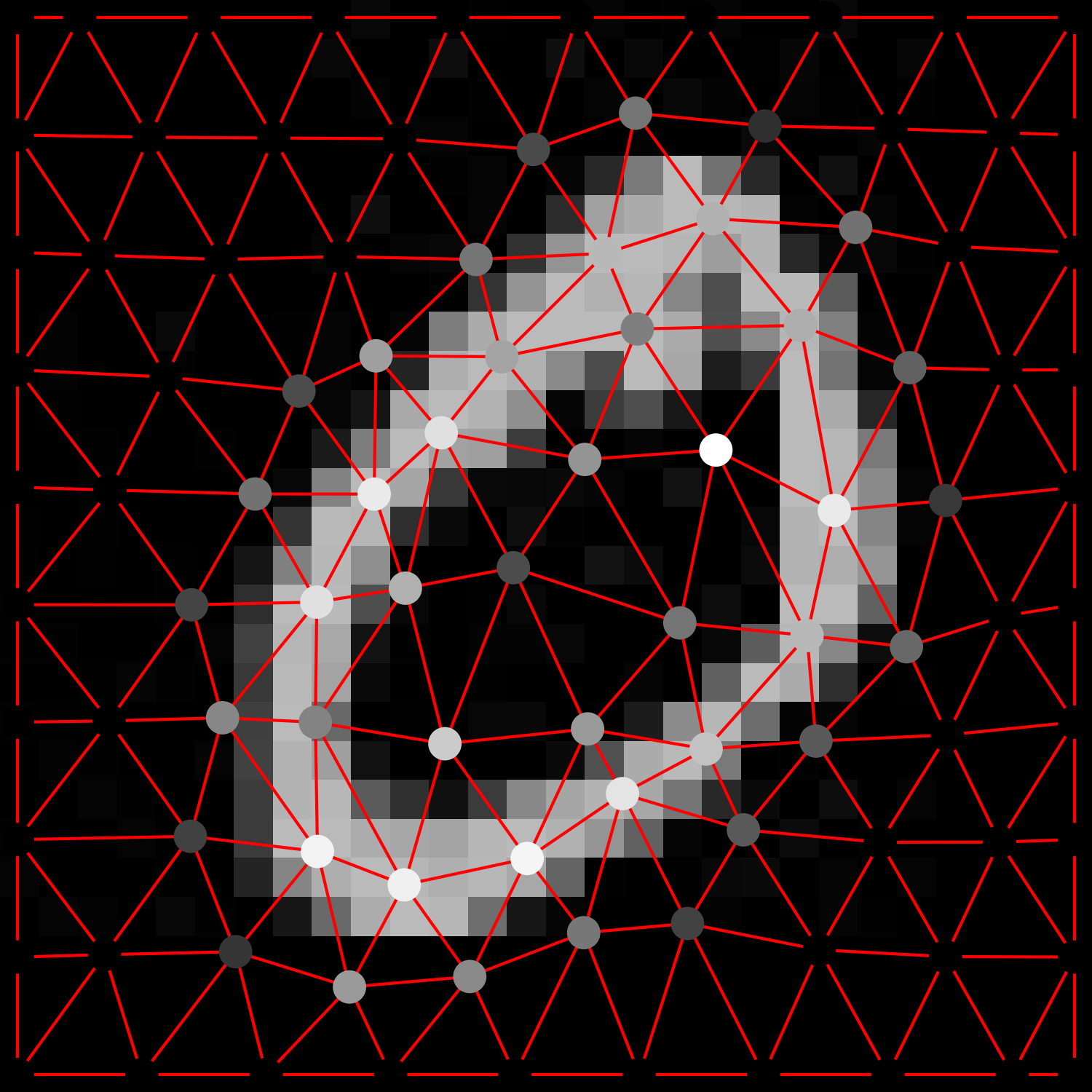}
    }
    \subfloat[AG $N=121$\label{fig:example_of_graphs_for_dgl:121:1}]{
	\includegraphics[width=0.15\linewidth]{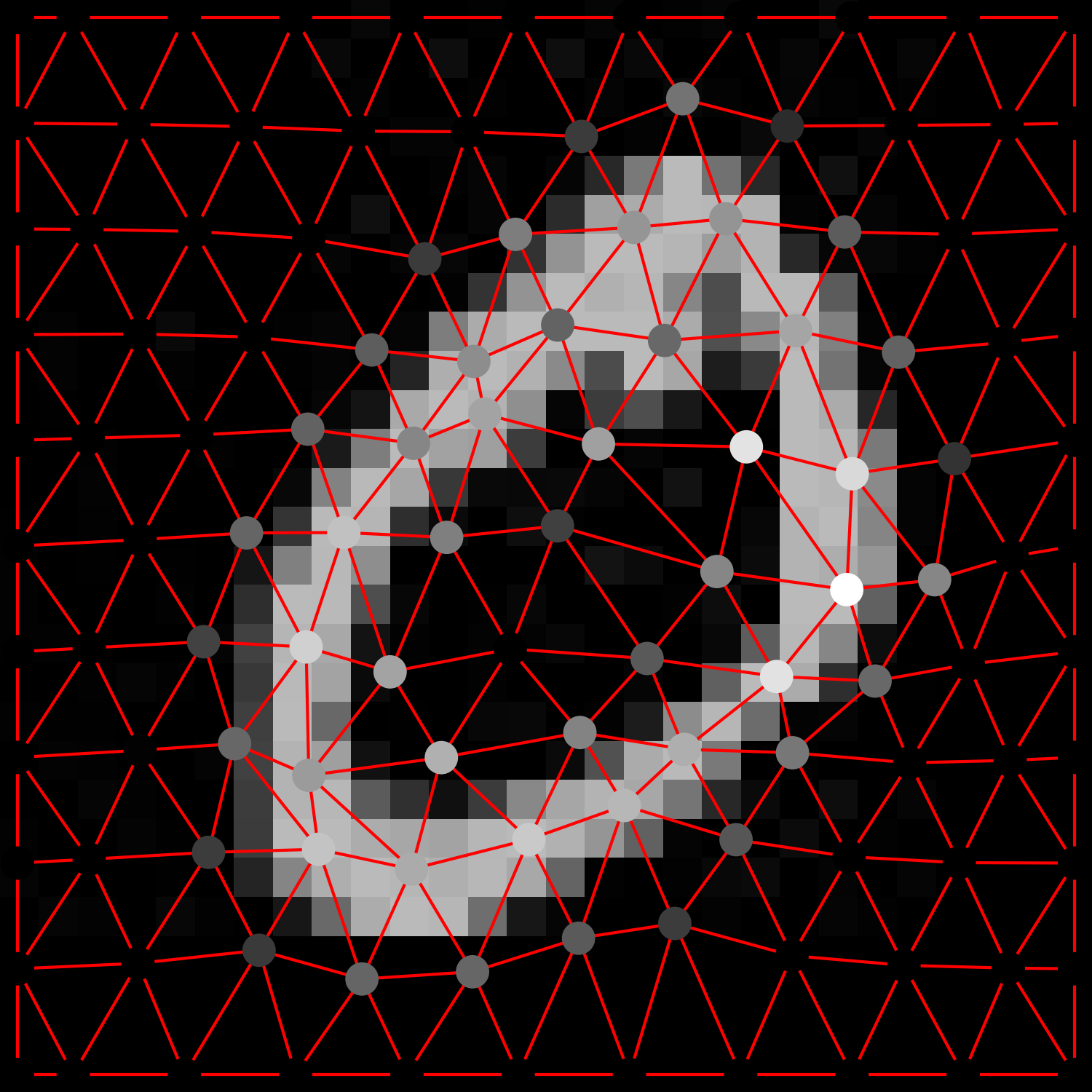}
    }
    \vspace{-1cm}
    \caption{Example adapted graphs derived from images in the MNIST dataset \citep{lecun1998gradient}. SP: superpixels, AG: adapted graph. The first row shows the nodes coloured by image intensity, the second row shows the nodes coloured by average saliency of the neighbouring edges. Edges are the same in both rows. \label{fig:example_of_graphs_for_dgl}} 
\end{figure*}

Examples of adapted graphs used in the geometric deep learning framework obtained from one MNIST image of a hand written '0' are shown in Fig. \ref{fig:example_of_graphs_for_dgl}. Edges are represented as red lines and nodes are represented by dots, with the grayscale indicating the magnitude of the node feature (white is higher value). The first row shows the graphs where the node feature is the intensity sampled from the image. In the second row, the node feature is the node saliency, computed as the average of the edge saliency for all over all edges connected to the node. 

\begin{figure}[htb!]
 	\centering
 	\includegraphics[width=\linewidth]{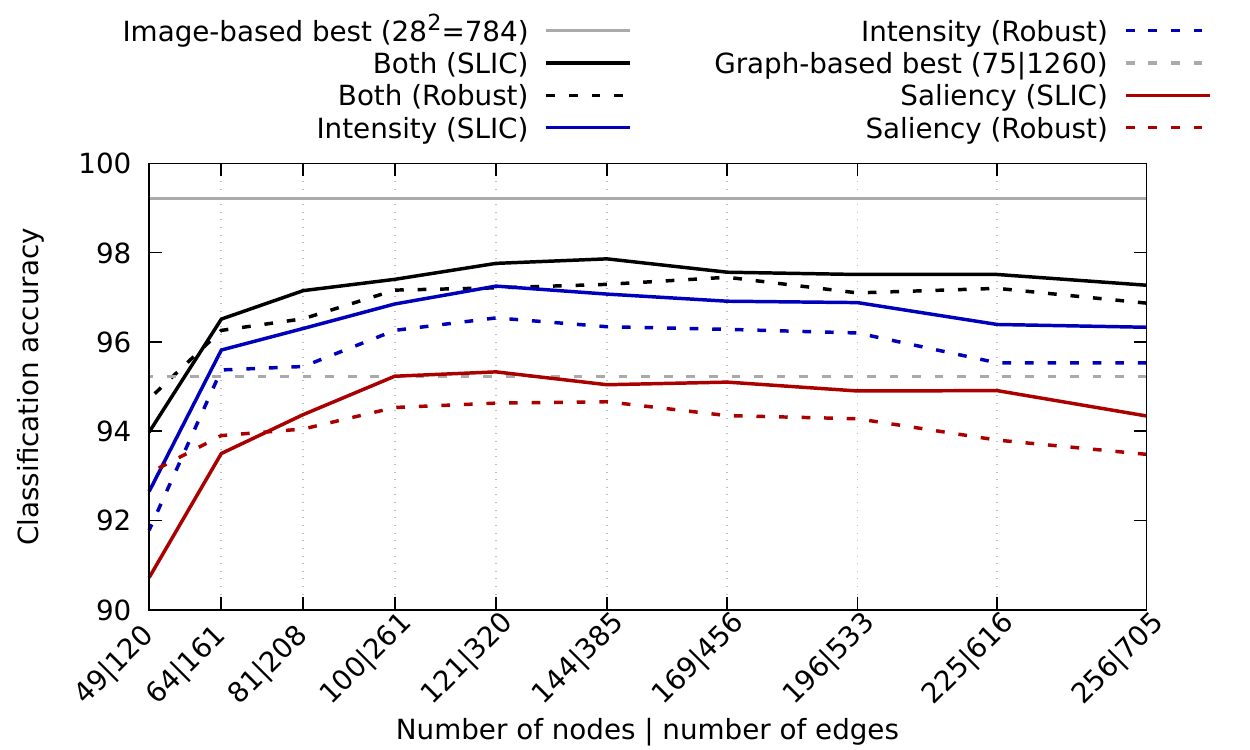}
     %
     \caption{Classification accuracy. Results are compared to the state of the art graph-based and image-based methods (in solid red horizontal lines) as reported by \cite{fey2018splinecnn}. \label{fig:classification_accuracy}} 
 \end{figure}

Figure \ref{fig:classification_accuracy} shows the results of the classification using SplineCNN from \cite{fey2018splinecnn}.  We present the classification results using the adapted graphs with both the robust saliency measure (solid lines) and the SLIC distance measure (dashed lines). For both cases, three types of features are considered: i) using only the intensity sampled from the images at node positions (which is equivalent to the superpixel approach in \cite{monti2017geometric,fey2018splinecnn}), ii) using only the proposed saliency measures, and iii) using both (two features per node). For each experiment we report not only the number of nodes of each graph, as done in related literature, but also the number of edges, which also influences the computational cost. 

Since the proposed method uses a uniform grid to initialise the graph and the images are square shaped, the number of nodes in our proposed graphs is always a squared value (49, 64, 81, ...). When comparing with previous work (using 75 superpixels) we draw the attention of the reader to the two closest cases: adapted graphs with 64 nodes, and with 81 nodes. The accuracy reported by \cite{fey2018splinecnn} in that case is 95.22\%. With our proposed graphs, we achieve a higher performance even with fewer nodes (64), using intensity only (95.37\% and 95.82\% using robust and SLIC saliency respectively) and clearly outperform competing graph based methods using both saliency and intensity features with only 64 nodes (96.26\% and 96.51\% using robust and SLIC saliency respectively). Higher number of nodes yield higher accuracy values, achieving up to 97.86\% with 144 nodes and 385 edges. This is to be put in context of state of the art classification results using the full grid (image) with LeNet \cite{lecun1998gradient} which has a reported accuracy of 99.33\% on the fully sampled images, closely followed by the work by \cite{fey2018splinecnn} where they reported 99.22\% on full images.

\section{Discussion and Conclusions}
\label{sec:discussion}

We have presented a novel method to extract graphs from images, that searches for salient image points  along edges, and the graph nodes are iteratively moved to the centroid of connected salient points. The method yields an adapted graph (edges cross feature lines in the images) and its dual (oversegmenting graph, with edges along image feature lines). The adaptation process lends itself naturally to a measurement of the edge saliency.

We have proposed two methods to find salient points along edges: a method inspired by SLIC superpixels \cite{Achanta2012}, and our new method which provides robustness against noise. Overall, the oversegmenting graphs that use the robust saliency measure were shown to outperform the SLIC-inspired (Distance-based) method for boundary adherence in presence of noise, achieving higher boundary recall and qualitatively better image adaptation. The difference, was less significant with natural images from the BSDS300 which are not particularly noisy.

We also computed adapted graphs on the MNIST dataset, to carry out digit classification using a geometric deep learning model --the SplineCNN by \cite{fey2018splinecnn}. Results were slightly better when using the SLIC distance saliency measure, probably because the images are not particularly noisy and in agreement with our results in boundary adherence. In both approaches we compared the effect of using, as node feature: sampled image intensity, saliency, and both. In the MNIST dataset, using intensity alone outperformed using saliency alone, possible for two reasons: first, MNIST images are grayscale and of high quality, hence bright pixels correlate very well with features, and dark pixels with background. As a result, image intensity is a good indicator of features. It is expected that other datasets will not adhere to this, and the role of saliency as a feature could be more prominent. Second, node saliency was computed as the average saliency of connected edges, since the algorithm provides saliency for edges, not for nodes. We decided to compute node saliency to enable direct comparison with intensity only and with superpixels, but other applications may benefit from edge features instead of node features. Using both features gave the best performance in all cases and outperformed the current state of the art results, achieved using graphs computed from a 75 superpixel partition of the images in \cite{fey2018splinecnn}.

Extension of the proposed method to 3D or higher dimensions is straightforward and requires no modification other than using a 3D graph initially. This is because all computations are carried out edge-wise or node-wise. Similarly, the proposed method has been described for scalar (grayscale) images, unlike most superpixel methods which are described for colour (RGB) images. Extension to colour images would only involve modifying the saliency point search. This is trivial for the SLIC distance (which is already described in the original SLIC paper \cite{Achanta2012}). Extension of the robust saliency measure to colour images could also be further investigated.

The proposed method was presented using an initial graph with uniform triangular topology, however the method is not limited by the type of initial graph. Because the adaptation process moves every graph node to the centroid of the salient points found on edges sharing the actually node, it is desirable (unless there is some problem-specific reason not to) that all nodes have the same number of connected edges so that every salient point contributes equally to moving all connected nodes. 

We have shown that the proposed graphs, with the associated saliency measure, can provide a relevant representation of shapes in images. Such representation may be useful in other computer vision problems that have not been covered here, for example image matching or feature detection, and more generally image processing methods that require a description of the topology of content in an image and where speed and computational efficiency are needed.

\section*{Acknowledgments}
 
This work was supported by the Wellcome Trust IEH Award  [102431]. The authors acknowledge financial support from the Department of Health via the National Institute for Health Research (NIHR) comprehensive Biomedical Research Centre award to Guy's \& St Thomas' NHS Foundation Trust in partnership with King's College London and King's College Hospital NHS Foundation Trust.

\section*{References}

\bibliography{mybib}
\bibliographystyle{model2-names}

\end{document}